\pdfoutput=1

\documentclass[11pt]{article}

\usepackage{acl}

\usepackage{times}
\usepackage{latexsym}
\usepackage{tabularx}
\usepackage{booktabs}
\usepackage[T1]{fontenc}

\usepackage[utf8]{inputenc}

\usepackage{microtype}

\usepackage{multirow}
\usepackage{booktabs}
\usepackage{amssymb,graphicx}
\usepackage{lipsum}
\usepackage{bm}
\usepackage{geometry}
\usepackage{booktabs}
\usepackage{multirow}
\usepackage{makecell}
\usepackage{latexsym}
\usepackage{booktabs}
\usepackage{diagbox}
\usepackage{url}
\usepackage{amsmath}
\usepackage{color}
\DeclareMathOperator*{\argmax}{argmax}

\usepackage{mathtools}%
\usepackage{bbm}
\usepackage{algorithm}
\usepackage{algorithmic}
\usepackage{latexsym}
\usepackage{multicol}
\usepackage{subfigure}
\usepackage{xspace}

\usepackage{enumitem}

\usepackage[normalem]{ulem}

\usepackage{todonotes}
\usepackage{amsmath}
\usepackage{cleveref}

\makeatletter
\newcommand*\iftodonotes{\if@todonotes@disabled\expandafter\@secondoftwo\else\expandafter\@firstoftwo\fi}  %
\makeatother

\newcommand{\moco}{\texttt{MoCo}\xspace}
\newcommand{\adcl}{\texttt{ADCL}\xspace}
\newcommand{\btcl}{\texttt{BTCL}\xspace}
\newcommand{\adv}{\texttt{ADV}\xspace}

\newcommand{\np}{\texttt{NP}\xspace}
\newcommand{\ftc}{\texttt{FTC}\xspace}
\newcommand{\ind}{\texttt{In-Domain}\xspace}
\newcommand{\out}{\texttt{Out-of-Domain}\xspace}

\newcount\Comments  %
\definecolor{darkgreen}{rgb}{0,0.5,0}
\definecolor{darkred}{rgb}{0.7,0,0}
\definecolor{teal}{rgb}{0.3,0.8,0.8}
\definecolor{blue}{rgb}{0,0,1}
\definecolor{purple}{rgb}{0.5,0,1}
\newcommand{\kibitz}[2]{\ifnum\Comments=1\textcolor{#1}{#2}\fi}

\title{Self-Supervised Contrastive Learning with Adversarial Perturbations for Defending Word Substitution-based Attacks}

\author{Zhao Meng\thanks{~~The first two authors contributed equally to this work.} \quad Yihan Dong$^*$ \quad Mrinmaya Sachan \quad \textbf{Roger Wattenhofer} \\
  ETH Zurich, Switzerland \\
    \texttt{\{zhmeng, yihdong, wattenhofer\}@ethz.ch} \\
    \texttt{mrinmaya.sachan@inf.ethz.ch}
  \\}

\begin{document}
\maketitle

\begin{abstract}
In this paper, we present an approach to improve the robustness of BERT language models against word substitution-based adversarial attacks by leveraging adversarial perturbations for self-supervised contrastive learning.
We create a word-level adversarial attack generating hard positives \textit{on-the-fly} as adversarial examples during contrastive learning.
In contrast to previous works, our method improves model robustness without using any labeled data. Experimental results show that our method improves robustness of BERT against four different word substitution-based adversarial attacks, and combining our method with adversarial training gives higher robustness than adversarial training alone.
As our method improves the robustness of BERT purely with unlabeled data, it opens up the possibility of using large text datasets to train robust language models against word substitution-based adversarial attacks.
\end{abstract}

\section{Introduction}
Pretrained language models such as BERT \cite[][inter alia]{bert} have had a tremendous impact on many NLP tasks. However, several researchers have demonstrated that these models are vulnerable to adversarial attacks, which fool the model by adding small perturbations to the model input~\cite{adv_machine_reading}. 

A prevailing method to improve model robustness against adversarial attacks is adversarial training~\cite{pgd}.
In NLP, adversarial training in the input space has been challenging, as existing natural language adversarial attacks are too slow to generate adversarial examples \textit{on the fly} during training~\cite{genetic, hotflip, pwws}. 
While some recent works~\cite{fgpm} have started exploring efficient input space adversarial training (e.g., for text classification), scaling adversarial training to pretrained language models like BERT has been challenging.

In this work, we in particular focus on improving the robustness of BERT against word substitution-based adversarial attacks. We propose an approach to adversarially finetune BERT-like models without using any labeled data. In order to achieve this, we rely on self-supervised contrastive learning~\cite{simclr}. Self-supervised contrastive learning has recently gained attention in the community and contrastive learning has been used to learn better representations for text classification~\cite{cl_sent, cl_sent2, simcse}. However, how to use these methods to improve model robustness remains an open question.

We combine self-supervised contrastive learning with adversarial perturbations by using adversarial attacks to generate hard positive examples for contrastive learning. To efficiently create adversarial examples, we leverage an adversarial attack, that is capable of generating multiple adversarial examples in parallel. The attack adversarially creates hard positive examples for contrastive learning by iteratively replacing words to follow the direction of the contrastive loss (see~\cref{fig:attack}). 

Experiments show that our method can improve the robustness of pretrained language models \textit{without looking at the labels} (in other words, before finetuning). Additionally, by combining our method with adversarial training, we are able to obtain better robustness than conducting adversarial training alone (see~\cref{sec:results}). 
Our study of the vector representations of clean examples and their corresponding adversarial examples indeed explains that our method improves model robustness by pulling clean examples and adversarial examples closer.

Our contributions\footnote{{We will release our code at \url{https://github.com/LotusDYH/ssl_robust}}} in this paper are two-fold. On the one hand, we improve the robustness of the pretrained language model BERT against word substitution-based adversarial attacks by using self-supervised contrastive learning with adversarial perturbations (see~\cref{sec:adv_cl}). On the other hand, to facilitate adversarial self-supervised contrastive learning, we create for BERT a word-level adversarial attack to create hard positive examples. The attack makes contrastive learning and adversarial training with \textit{on-the-fly} generated adversarial examples possible.
Additionally, we also show that our method is capable of using out-of-domain data to improve model robustness (see~\cref{tab:ood} and~\cref{sec:results}). This opens an opportunity for using large-scale unlabeled data to train robust language models against word substitution-based adversarial attacks.

\begin{figure*}
    \centering
    \includegraphics[width=0.75\linewidth]{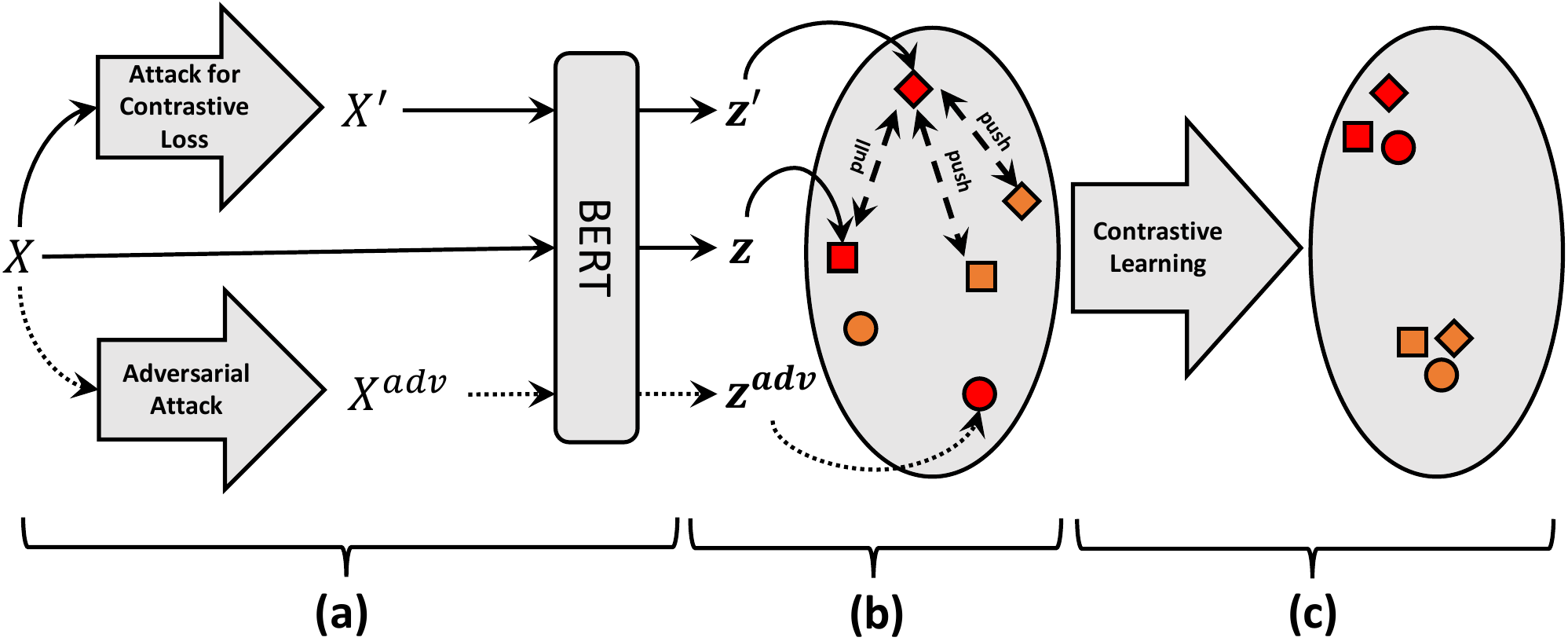}

    \caption{An illustration of our method. \textbf{(a)} For the original example $X$, we obtain the hard positive example $X^\prime$ by Geometry Attack for contrastive loss (see~\cref{sec:attack}). \textbf{(b)} Before contrastive learning, in the vector space, the clean example $\bm z$, the hard positive example $\bm z^\prime$, and the adversarial example $\bm z^{adv}$ are far from each other. Contrastive learning pulls the clean example $\bm z$, and the hard positive example $\bm z^\prime$ together. \textbf{(c)} After contrastive learning, the clean example $\bm z$, the hard positive example $\bm z^\prime$, and the adversarial example $\bm z^{adv}$ are close. We omit MLP in this figure for simplicity. We use a different color to show another example from the dataset. See~\cref{sec:method} for details. Note that the adversarial example $X^{adv}$ and its corresponding vector $\bm z^{adv}$ are not used in contrastive learning. We nevertheless show $X^{adv}$ and $\bm z^{adv}$ for illustration purposes.}
    \label{fig:method}
\end{figure*}
\section{Related Work}

\subsection{Adversarial Training for NLP}\label{sec:related_adv}

Adversarial training improves model robustness by augmenting clean examples with adversarial examples during training.
Previous works on adversarial training for natural language mainly focus on perturbations in the vector space, while actual adversarial attacks create adversarial examples by changing natural language symbols. For example,~\citet{freelb} and~\citet{alum} improve model generalization ability by adversarial training on the word embedding space, without mentioning model robustness. However, they either ignore model robustness, or only test model robustness against the adversarial dataset ANLI, without paying attention to actual adversarial attacks. Other works conduct adversarial training in the word space~\cite{genetic, pwws}. Still, they can only do adversarial training on a limited number of pre-generated adversarial examples due to the low efficiency of the attacks.~A recent work~\cite{fgpm} conducts adversarial training efficiently in the word space, but their method is limited to non-contextualized models. 

Apart from adversarial training, other supervised learning methods~\cite{ascc, dne, infobert, tavat} have also been proposed to improve robustness. However, these methods are supervised and are not comparable to our work.

Our work also differs from previous works in natural language adversarial training. On the one hand, as opposed to previous works, which are supervised, we propose a self-supervised learning scheme to improve the robustness of pretrained language models. On the other hand, while previous works mostly focus on adversarial training in embedding space, we conduct efficient adversarial training with pretrained language models at the word level. 

\subsection{Contrastive Learning for NLP}\label{sec:related_cl}

Contrastive learning was first proposed in the image domain to improve model performance in a self-supervised fashion~\cite{moco, simclr}. These methods bring representations of similar examples closer and push representations of dissimilar examples further apart. Additionally, researchers also find that by adding adversarial perturbations during contrastive learning, image classification models become more robust against adversarial attacks~\cite{rocl}.

In NLP, previous works on contrastive learning mainly focus on improving model generalization.~\citet{scl_nlp} boost performance of RoBERTa by adding supervised signals during finetuning on downstream tasks. ~\citet{cl_gen} tackle the ``exposure bias" problem in text generation by adding adversarial signals during contrastive learning. Other similar works include~\citet{cl_mt},~\citet{cl_sent}, and~\citet{simcse}.
Although these works have demonstrated the usefulness of contrastive learning in NLP applications, few have addressed the robustness of NLP models, particularly pretrained language models, against word substitution-based natural language adversarial attacks.  

Recently,~\citet{cline} claimed that their method improves model robustness against adversarial sets. However, such sets are pre-generated and are less challenging than adversarial examples generated on the fly by actual adversarial attacks~\cite{textfooler, pwws}. 
In this paper, we focus on improving the robustness of pretrained language models against word substitution-based adversarial attacks. We present the details of our method in~\cref{sec:method}.

\section{Methodology}\label{sec:method}

In this section, we describe our method for self-supervised contrastive learning with adversarial perturbations. Specifically, \cref{sec:motivation} gives the background and motivation of our problem, and \cref{sec:adv_cl} describes the adversarial contrastive learning framework. Finally, in~\cref {sec:attack}, we describe the adversarial attack used in contrastive learning.

\subsection{Background and Motivation}\label{sec:motivation}
In this work, we focus on text classification tasks\footnote{Although our formulation can also be extended to several other problems.}.
Let us assume that we have an example text $X_i = \{w_1, w_2, \dots, w_L\}$ with $L$ words and let $y_i$ be the corresponding class label for $X_i$. Our text classification model consists of a BERT encoder $f(\cdot)$ and an MLP classification head $c(\cdot)$.

We obtain the vector representation $\bm h_i \in \mathbb{R}^d$ of the example $X_i$ by feeding $X_i$ into the BERT encoder $f(\cdot)$. Then the MLP classification head $c(\cdot)$ takes $\bm h_i$ as input to give us the prediction. Formally, we have: 

\begin{align*}
    \bm h_i &= f(X_i) \\
    \hat{y}_i &= c(\bm h_i)
\end{align*}

\noindent where $\hat{y}_i$ is the predicted label. We have $\hat{y}_i = y_i$ if the model prediction is correct. 

A word substitution-based adversarial attack $a(\cdot)$ takes an original example $X_i$ as input and generates an adversarial example $X_i^{adv}$ by substituting the $k$-th original word $w_k$ in $X_i$ with another word $w_k^{adv}$. To make the orignal example $X_i$ and the adversarial example $X_i^{adv}$ close in semantics, existing works often use synonyms as substitutions~\cite{pwws, textattack}. 

By conducting the word substitution, the attack $a(\cdot)$ aims to fool the model with $X_i^{adv}$. Formally, we have:

\begin{align*}
    X_i^{adv} &= a(X_i) \\
    \bm h_i^{adv} &= f(X_i^{adv}) \\
    \hat{y}_i^{adv} &= c(\bm h_i^{adv})
\end{align*}

\noindent where $X_i^{adv} = \{w_1, w_2, \dots, w_k^{adv}, \dots, w_L  \}, 1 \leq k \leq L$. Assuming the attack successfully fools the model, we would have $\hat{y}_i \neq \hat{y}_i^{adv}$. The key assumption in our approach is that although $X_i$ and $X_i^{adv}$ are very similar to each other at the word level, it is possible that the encoder $f$ embeds them in such a way that the distance between their representations $\bm h_i$ and $\bm h_i^{adv}$ are large and the classification head $c(\cdot)$ predicts $X_i$ and $X_i^{adv}$ to be of different classes. 

Thus, the goal of our method is to obtain a robust model, on which we have $y_i = \hat{y}_i$ and $\hat{y}_i = y_i^{adv}$. In other words, the robust model defends an adversarial example $X_i^{adv}$ of the original example $X_i$ successfully, if the robust model gives the same correct prediction on the original example $X_i$ and the adversarial example $X_i^{adv}$. We use \textit{attack success rate} as the evaluation metric for model robustness. The attack success rate is defined as the rate of an attack successfully fooling the model on all test examples. 

To obtain a robust model, we optimize the encoder such that $\bm h_i$ and $\bm h_i^{adv}$ become similar to each other. We achieve this goal by conducting self-supervised contrastive learning on the encoder with adversarial perturbations, during which we use an attack to create \textit{hard positive examples}, maximizing the contrastive loss. The rest this section gives the details of our method.

\subsection{Self-Supervised Contrastive Learning with Adversarial Perturbations}\label{sec:adv_cl}

Following previous works on self-supervised contrastive learning~\cite{moco, simclr}, we formulate our learning objectives as follows.
Consider we have a batch of $n$ examples and $X_i$ is the $i$-th input, we first obtain $X_i^\prime = t(X_i)$ as an augmentation of $X_i$ by transformation $t(\cdot)$.
We call $X_i$ and $X_i^\prime$ a pair of positive examples. All other examples in the same batch are considered negative examples of $X_i$ and $X_i^\prime$. 

To take advantage from using more negative examples, we use \moco~\cite{moco} as our framework, in which we employ an encoder $f_q$ for the positive examples, and another momentum encoder $f_k$ for the negative examples. We then have:
\begin{align*}
    \bm h_i &= f_q(X_i) \\
    \bm h_i^\prime &= f_k(X_i^\prime)
\end{align*}

\noindent where $\bm h_i, \bm h_i^\prime \in \mathbb{R}^d$ are representations of $X_i$ and $X_i^\prime$, respectively. 
During training, $f_q$ and $f_k$ are initialized the same. We update $f_k$ momentarily: 

\begin{align*}
    \theta_k \gets m\cdot \theta_k + (1-m) \cdot \theta_q
\end{align*}

\noindent where $\theta_k$ and $\theta_q$ denote the parameters of $f_k$ and $f_q$, respectively. We then have:

\begin{align*}
    \bm z_i &= g_q(\bm h_i) \\
    \bm z_i^\prime &= g_q(\bm h_i^\prime)
\end{align*}

\noindent where $\bm z_i, \bm z_i^\prime \in \mathbb{R}^c$, $g_q(\cdot)$ and $g_k(\cdot)$ are MLPs with one hidden layer of sigmoid activation, respectively. Following~\citet{simclr}, we conduct contrastive learning on $\bm z$ instead of $\bm h$ to prevent the contrastive learning objective from removing information useful for downstream tasks. After contrastive learning, we use $\bm h$ as the sentence representation for downstream tasks.

Additionally, we also maintain a dynamic first-in-first-out queue for the negative examples. During training, before computing contrastive loss at the end of each batch, all encoded examples of the current batch are enqueued into the queue, and the oldest examples are dequeued simultaneously.

In our experiments, we use the attack described in~\cref{sec:attack} or back-translation~\cite{bk} for augmentation $t(\cdot)$.  
Assume that we have an encoded example $\bm z_i$ and the encoded examples in the queue are \{$\bm z_0, \bm z_1, \cdots, \bm z_{Q-1}$\}, where $Q$ is the size of the queue. Among the encoded examples in the queue, one of them is $\bm z_i^\prime$, which forms a pair of positive examples with $\bm z_i$. 
We use contrastive loss to maximize the similarity between positive examples, while minimizing the similarity of negative examples. 
We then have:
\begin{equation}
\label{eq:loss}
    \ell_{i} = -\log \frac{\exp(\mathrm{sim}(\bm z_i, \bm z_i^\prime)/\tau)}{\sum_{k=0}^{Q}\exp(\mathrm{sim}(\bm z_i, \bm z_{k})/\tau)}
\end{equation}

\noindent where $\tau$ is the temperature parameter, $\mathrm{sim}(\cdot, \cdot)$ is the similarity function, and $Q$ is the size of the dynamic queue. In this paper,  we compute similarity by dot product as in \moco.

By optimizing~\cref{eq:loss}, the goal is to maximize the similarity of representations between similar (positive) pairs of examples while minimizing the similarity of representations between dissimilar (negative) examples. We use the geometry-inspired attack described in~\cref{sec:attack} as the transformation $t(\cdot)$ to create pairs of examples that are similar on the word level but at the same time are distant from each other in the representation space. 

We illustrate our method in~\cref{fig:method}. In~\cref{fig:method} (b) and (c), by conducting contrastive learning and using the Geometry Attack generated adversarial examples as hard positives, the vector representations obtained from the model become \textit{invariant} to the adversarial attacks.

\begin{figure}[!ht]
    \centering
    \includegraphics[width=0.5\linewidth]{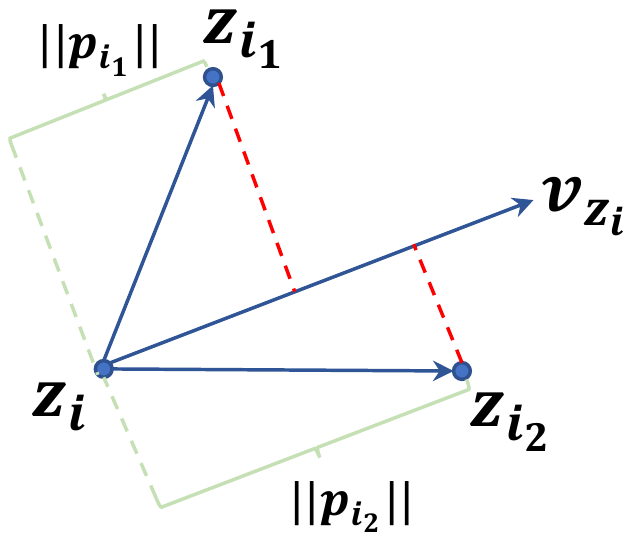}

    \caption{An illustration of one iteration in Geometry Attack for contrastive loss. See~\cref{sec:attack} for details.}
    \label{fig:attack}
\end{figure}
\subsection{Creating Hard Positive Examples by Geometry Attack}\label{sec:attack}

As mentioned in \cref{sec:adv_cl}, we use an attack as the transformation $t(\cdot)$ during contrastive learning. We describe how this attack creates adversarial examples for contrastive loss during self-supervised contrastive learning (see~\cref{fig:method} (b)) in this subsection.

Inspired by~\citet{geometry}, who leverage geometry of representations to generate natural language adversarial examples for text classification tasks, we also use the geometry of pretrained representations to create adversarial examples for contrastive loss. 
The created adversarial examples are used as positive examples of the original examples in our contrastive learning framework, and at the same time are created to maximize the contrastive loss. Hence, we refer to adversarial examples created by the attack as \textit{hard positive examples}.

The intuition of our attack is that we repeatedly replace words in the original texts such that in each iteration, the replaced word increases the contrastive loss as much as possible. To be specific, consider an example $X_i$, we then have:

\begin{itemize}[topsep=0pt,itemsep=0pt,partopsep=0pt, parsep=0pt, leftmargin=*] 
    \item[1.] \textbf{Determine Direction for Sentence Vector} Compute the gradients of $\ell_i$ with respect to $\bm z_i$. In this step, we find the direction we should move from $\bm z_i$ to increase the contrastive loss $\ell_i$. We have the gradient vector $\bm v_{z_i} = \nabla_{\bm z_i} \ell_i$.
    
    \item[2.] \textbf{Choose Original Word to be Replaced} Compute the gradients of $\ell_i$ with respect to input word embeddings of $X_i$. For words tokenized into multiple tokens, we take the average of the gradients of the tokens. In this step, we find the word $w_t$ which has the most influence in computing $\ell_i$. Specifically, assuming we have $L$ words, then we choose $t = \arg\max_t \{ ||\bm g_1||, ||\bm g_2||, \dots, ||\bm g_L|| \}$, where $\bm g_k$ is the gradients of $l_i$ with respect to the embeddings of word $w_k$, $1 \leq k \leq L$.

    \item[3.] \textbf{Generate Candidate Set} Suppose we choose the word $w_t$ in step 2. In this step, we use a pretrained BERT to choose the most probable candidates $w_t$ to replace it in the original text. We have the candidates set = $\{w_{t_1}, w_{t_2}, \cdots, w_{t_T}\}$. Following~\citet{textfooler}, we filter out semantically different words from the candidate set by discarding candidate words of which the cosine similarity of their embeddings between the embeddings of $w_t$ is below a threshold $\epsilon$. We set the threshold $\epsilon = 0.5$ and use counter-fitted word embeddings~\cite{counter} to compute the cosine similarity.
    
    \item[4.] \textbf{Choose Replacement Word} Replace $w_t$ with words in the candidates set, resulting in text vectors $\{\bm z_{i_1}, \bm z_{i_2}, \cdots,\bm z_{i_T} \}$. We compute delta vector $\bm r_{i_j} \gets \bm z_{i_j} - \bm z_i$. The projection of $\bm r_{i_j}$ onto $\bm v_{z_i}$ is: $\bm p_{i_j} \gets \frac{\bm r_{i_j} \cdot \bm v_{z_i}}{||\bm v_{z_i}||}$. We select word $w_{t_m}$, where $m \gets \argmax_j  ||\bm p_{i_j}||$. In other words, $w_{t_m}$ results in the largest projection $\bm p_{i_m}$ onto $\bm v_{z_i}$.
    
    \item[5.] \textbf{Repetition} Replace $w_t$ with $w_{t_m}$ in $X_i$, then we have $\bm z_i \gets \bm z_{i_m} $. Repeat step 1-4 for $N$ iterations, where $N$ is a hyperparameter of our method. We expect $\ell_i$ to increase in each iteration.
\end{itemize}

\Cref{fig:attack} illustrates an iteration of our attack, in which we have two options to choose from the candidate set. This attack can be easily implemented in a batched fashion, making it possible for us to generate adversarial examples \textit{on the fly} during training. Furthermore, our efficient implementation makes it possible to conduct contrastive learning with adversarial perturbations as well as adversarial training with adversarial examples generated on the fly. We give a speed comparison of our attack and other attacks in ~\cref{app:speed}. We also give pseudocode of the attack in~\cref{alg:attack} of~\cref{app:geo}.

\section{Experiments}

\subsection{Datasets and Evaluation Metrics}
We test how our method improves model robustness on four text classification datasets: AG's News, Yelp, IMDB, and DBpedia (See~\cref{app:dataset} for details).

We report the \textit{attack success rate} and the \textit{replacement rate} of the attacks as the evaluation metrics. Following~\citet{genetic, hotflip}, to prevent the model accuracy on clean examples from confounding the results, we define the success rate of an attack on all \textit{correctly classified} examples in the test set. Lower success rates indicate higher robustness. The replacement rate refers to the percent of original words replaced in the clean example. Higher replacement rates indicate that the attack needs to replace more words to fool the model, and thus mean that the model is more robust.

\subsection{Attacks for Evaluating Robustness}\label{sec:eval_attacks}

We use four word substitution-based adversarial attacks to evaluate the model robustness. 

\noindent\textbf{Geometry Attack} We use the same attack described in~\cref{sec:attack} to generate adversarial examples for sentence classification tasks by replacing contrastive loss with \textit{cross-entropy classification loss}. We set the maximum number of replaced words to 20. 

\noindent\textbf{TextFooler}, \textbf{PWWS}, and \textbf{BAE-R} We use the default implementations from TextAttack~\cite{textattack}.

All these attacks will give up and terminate once the \textit{maximum number of replaced words} (sometimes also called \textit{perturbation budget}) is reached.

\subsection{Experimental Design}\label{sec:exp_designs}

We have the following hypotheses for our method:

\noindent\fbox{%
    \parbox{\linewidth}{%
\noindent {\textbf{H1}}: Self-supervised contrastive learning improves model robustness against adversarial attacks. Moreover, using adversarial perturbations during contrastive learning further improves robustness.
    }%
}

\noindent To validate this hypothesis, we set three different pretraining schemes:

\noindent {\btcl}: Pretraining with back-translation as the transformation $t(\cdot)$ for self-supervised contrastive learning.

\noindent {\adcl}: Pretraining with Geometry Attack for contrastive loss (see~\cref{sec:attack}) as transformation $t(\cdot)$ for self-supervised contrastive learning. 

\noindent {\np}: Apart from the above two settings, we also add a \textit{No Pretraining} baseline to understand the general effectiveness of contrastive learning. 

\noindent\fbox{%
    \parbox{\linewidth}{%
\noindent {\textbf{H2}}: Combining self-supervised contrastive learning with adversarial training gives higher robustness than conducting adversarial training alone.
    }%
}

\noindent We use different finetune strategies to understand how adding adversarial training to our method affects model robustness. We have two settings:

\noindent {\ftc}: We finetune the pretrained model on the clean examples of the corresponding downstream dataset.

\noindent {\adv}: We conduct adversarial training by leveraging supervisedly generated adversarial examples. Note that our adversarial training is different from previous works~\cite{pwws, genetic}, which merely finetune the model on a fixed number of pre-generated adversarial examples. Instead, our adversarial training scheme is similar to~\citet{pgd}, where the model is finetuned on clean examples and adversarial examples generated \textit{on the fly} during each batch of training. 

We use Geometry Attack for adversarial training as the remaining three attacks are not efficient enough to generate adversarial examples \textit{on the fly} (see~\cref{app:speed} for details).

\noindent\fbox{%
    \parbox{\linewidth}{%
\noindent {\textbf{H3}}: Our contrastive learning method is capable of using out-of-domain data to improve the model robustness.
    }%
}

\noindent While in H1 and H2, we use the same dataset for pretraining and finetuning, we want to test how our method can leverage out-of-domain data. Hence, we have two additional experimental settings:

\noindent \ind: We use the same dataset during contrastive learning and finetuning. 

\noindent \out: We use different datasets for contrastive learning and finetuning.

\noindent\fbox{%
    \parbox{\linewidth}{%
\noindent {\textbf{H4}}: By optimizing~\cref{eq:loss}, our method pulls the representations of the clean samples and their corresponding hard positive examples closer in the vector space while pushing other different examples further. In this way, the representations of clean examples and their adversarial examples are also closer in the vector space.
    }%
}

\noindent We validate this hypothesis by conducting a vector space study. See~\cref{sec:results} for details.

Note that to avoid confusing adversarial examples generated during contrastive learning and adversarial examples generated during finetuning, we refer to the former as \textit{hard positive examples} (see~\cref{sec:attack}). 

\begin{table*}[!t]
\centering
\resizebox{\linewidth}{!}{%

\begin{tabular}{cccc>{\centering\arraybackslash}p{1.4cm}>{\centering\arraybackslash}p{1.4cm}>{\centering\arraybackslash}p{1.4cm}>{\centering\arraybackslash}p{1.4cm}>{\centering\arraybackslash}p{1.4cm}>{\centering\arraybackslash}p{1.4cm}>{\centering\arraybackslash}p{1.4cm}>{\centering\arraybackslash}p{1.4cm}}
\toprule
\multirow{2}*{\textbf{Dataset}} & \multirow{2}*{\textbf{Pretrain}} &  \multirow{2}*{\textbf{Finetune}} & \multirow{2}*{\textbf{ Acc. (\%)}} & \multicolumn{4}{c}{\textbf{Success Rate (\%) $\downarrow$}} & \multicolumn{4}{c}{\textbf{Replaced (\%) $\uparrow$}}\\ 

\cmidrule(lr){5-8} \cmidrule(lr){9-12} & & & & Geometry & TextFooler & PWWS & BAE-R  & Geometry & TextFooler & PWWS & BAE-R\\
\midrule
\midrule

\multirow{5}*{\textbf{AG}} & \multirow{2}*{\np} & \ftc & $94.2$ & $86.2 $ & $87.6 $ & $63.6 $ & $17.9 $ & $18.6$ & $25.7$ & $20.9$ & $7.4$
\\   & & \adv & $94.4$ & $\textit{20.7}$ & $\textit{25.1}$ & $\textit{26.1}$ & $\textit{10.7}$ & $\textit{20.5}$ & $\textit{29.3}$ & $\textbf{22.3}$ & $\textbf{7.7}$
\\\cmidrule(lr){2-3} \cmidrule(lr){4-4} \cmidrule(lr){5-8} \cmidrule(lr){9-12}
   & \btcl & \ftc & $94.4$ & $80.6$ & $84.6$ & $63.1$ & $17.7$  & $18.1$ & $24.6$ & $20.9$ & $7.5$ 
\\\cmidrule(lr){2-3} \cmidrule(lr){4-4} \cmidrule(lr){5-8} \cmidrule(lr){9-12}
  & \multirow{2}*{\adcl} & \ftc & $94.3$ & ${76.5}$ & ${80.7}$ & ${55.9}$ & ${14.1}$ & $19.1$ & $26.7$ & $\textit{22.6}$ & $\textit{7.5}$ 
\\   & & \adv & $94.4$ & $\textbf{18.7}$ & $\textbf{23.5}$ & $\textbf{24.7}$ & $\textbf{9.7}$ & $\textbf{20.6}$ & $\textbf{
29.3}$ & $22.2$ & $7.2$ \\
\midrule

\multirow{5}*{\textbf{Yelp}} & \multirow{2}*{\np} & \ftc & $97.1$ & $94.6$ & $94.3$ & $97.0$ & $42.1$ & $10.6$ & $10.4$ & $7.1$  & $6.7$
\\   & & \adv & $96.2$ & $\textit{38.8}$ & $\textit{52.4}$ & $\textit{62.7}$ & $\textit{22.2}$ & $\textit{12.8}$ & $\textbf{17.3}$ & $\textbf{11.3}$ & $\textbf{8.8}$ 
\\\cmidrule(lr){2-3} \cmidrule(lr){4-4} \cmidrule(lr){5-8} \cmidrule(lr){9-12}
   & \btcl & \ftc & $97.1$ & $92.3$ & $91.6$ & $94.8$ & $39.2$ & $11.0$ & $10.1$ & $7.7$ & $6.9$ 
\\\cmidrule(lr){2-3} \cmidrule(lr){4-4} \cmidrule(lr){5-8} \cmidrule(lr){9-12}
  & \multirow{2}*{\adcl} & \ftc & $97.0$ & ${88.6}$ & ${88.2}$ & ${91.1}$ & ${37.8}$ & $10.4$ & $10.5$ & $7.4$ & $6.9$
\\   & & \adv & $96.1$ & $\textbf{35.6}$ & $\textbf{50.1}$ & $\textbf{61.0}$ & $\textbf{21.0}$ & $\textbf{13.4}$  & $\textit{17.1}$ & $\textit{11.2}$ & $\textit{8.3}$ \\
\midrule

\multirow{5}*{\textbf{IMDB}} & \multirow{2}*{\np} & \ftc & $92.3$ & $98.7$ & $99.0$ & $99.2$ & $54.0$ & $3.5$ & $6.5$ & $4.3$ & $3.0$
\\   & & \adv & $92.0$ & $\textit{51.4}$ & $\textit{75.3}$ & $\textit{79.1}$ & $\textit{35.1}$ & $\textit{7.4}$ & $\textbf{12.7}$ & $\textbf{9.3}$ & $\textbf{3.6}$ 
\\\cmidrule(lr){2-3} \cmidrule(lr){4-4} \cmidrule(lr){5-8} \cmidrule(lr){9-12}
   & \btcl & \ftc & $92.5$ & $93.3$ & $96.6$ & $95.1$ & $52.0$ & $4.5$ & $7.4$ & $4.4$ & $3.3$
\\\cmidrule(lr){2-3} \cmidrule(lr){4-4} \cmidrule(lr){5-8} \cmidrule(lr){9-12}
  & \multirow{2}*{\adcl} & \ftc & $92.4$ & ${84.2}$ & ${87.8}$ & ${87.8}$ & ${48.0}$ & $3.7$ & $8.7$ & $5.1$ & $2.3$
\\   & & \adv & $91.9$ & $\textbf{48.7}$ & $\textbf{74.4}$ & $\textbf{77.6}$ & $\textbf{31.8}$ & $\textbf{8.1}$  & $\textit{12.4}$ & $\textit{9.1}$ & $\textit{3.5}$ \\
\midrule

\multirow{5}*{\textbf{DBpedia}} & \multirow{2}*{\np} & \ftc & $99.2$ & $79.6$ & $79.3$ & $46.7$ & $14.3$ & $17.8$ & $23.2$ & $16.2$ & $13.3$
\\   & & \adv & $99.0$ & $\textit{13.9}$ & $\textit{16.5}$ & $\textit{17.7}$ & $\textit{10.9}$ & $\textbf{21.6}$ & $\textit{28.2}$ & $\textbf{18.9}$ & $\textbf{14.1}$
\\\cmidrule(lr){2-3} \cmidrule(lr){4-4} \cmidrule(lr){5-8} \cmidrule(lr){9-12}
   & \btcl & \ftc & $99.2$ & $77.4$ & $76.8$ & $45.1$ & $13.0$ & $18.9$ & $22.8$ & $18.1$ & $13.1$
\\\cmidrule(lr){2-3} \cmidrule(lr){4-4} \cmidrule(lr){5-8} \cmidrule(lr){9-12}
  & \multirow{2}*{\adcl} & \ftc & 99.1 & ${73.6}$ & ${74.5}$ & ${42.6}$ & ${11.6}$ & $18.2$ & $22.9$ & $17.6$ & $12.8$
\\   & & \adv & 99.0 & $\textbf{12.4}$ & $\textbf{14.8}$ & $\textbf{16.2}$ & $\textbf{10.1}$ & $\textit{20.1}$ & $\textbf{28.6}$ & $\textit{18.2}$ & $\textit{13.8}$ \\

 \bottomrule
\end{tabular}
}

\caption{\label{tab:white-box}
Experimental results for H1 and H2. \ind setting is used. We \textbf{bold} the best results, while the second best is in \textit{italic}.
}
\end{table*}

\subsection{Results}\label{sec:results}

Table~\ref{tab:white-box} shows the experimental results for validating H1 and H2. For each dataset, when evaluating the model robustness, we use the same perturbation budget across different settings. Note that although the replacement rates vary across different settings of the same dataset, \textit{the perturbation budget for the same attack is the same} in these settings. By using the same perturbation budget, we ensure that the success rates of the attacks provide us with a fair evaluation of the robustness of the model~\cite{fgpm, pwws}.

\noindent \textbf{H1}: 
To validate H1, we focus on rows with the \ftc setting during finetuning. We can observe that models without any contrastive pretraining (\np) are the most vulnerable to adversarial attacks. For example, the success rate of the Geometry Attack for AG's News dataset is 86.2\% for the \np model. In contrast, for \btcl and \adcl, the success rate of the Geometry Attack is at least 5.6\% lower than this setting. This shows that \textit{self-supervised contrastive learning does improve model robustness}. 

Additionally, we can also see from \cref{tab:white-box} that \adcl improves the model robustness more than \btcl. For example, in the IMDB dataset, the model pretrained with \adcl is 9.1\% more robust than the model pretrained with \btcl ($93.3\% \rightarrow 84.2\%$), showing that \textit{using adversarial perturbations during contrastive learning further improves model robustness against adversarial attacks}. Hence, we do not combine \btcl with \adv in later experiments for simplicity.

To further understand how contrastive learning improves the model robustness, we study the transferability of the adversarial examples between models without any contrastive pretraining (\np) and the models pretrained with \adcl. To be specific, the models are first pretrained using either \np or \adcl, and then finetuned on clean examples (\ftc). Then, we use a \np model to generate adversarial examples on the test set of each dataset, and then test the corresponding model pretrained with \adcl on these adversarial examples. And vice versa. 

Table~\ref{tab:transfer} shows the results. We can see that adversarial examples generated by models pretrained with \adcl have much higher success rates on models without any contrastive pretraining (\np). For example, for the AG's News dataset, the success rates increase by 32.1\%, 35.3\%, 33.8\%, and 22.1\% for Geometry Attack, TextFooler, BAE-R, and PWWS, respectively. This demonstrates that by self-supervised contrastive learning with adversarial perturbations, the models become more robust against attacks. 

\begin{table*}
\renewcommand\arraystretch{1.0}
\centering
\resizebox{\linewidth}{!}{%

\begin{tabular}{cccc>{\centering\arraybackslash}p{1.3cm}>{\centering\arraybackslash}p{1.3cm}>{\centering\arraybackslash}p{1.3cm}>{\centering\arraybackslash}p{1.3cm}>{\centering\arraybackslash}p{1.3cm}>{\centering\arraybackslash}p{1.3cm}>{\centering\arraybackslash}p{1.3cm}>{\centering\arraybackslash}p{1.3cm}}
\toprule
\multirow{2}*{\textbf{Dataset}} & \multirow{2}*{\textbf{Domain}} &\multirow{2}*{\textbf{Pretrain}} & \multirow{2}*{\textbf{Acc.~(\%)}} & \multicolumn{4}{c}{\textbf{Success Rate (\%) $\downarrow$}} & \multicolumn{4}{c}{\textbf{Replaced~(\%)$\uparrow$}}\\ 
\cmidrule(lr){5-8} \cmidrule(lr){9-12} & & & & Geometry & TextFooler & PWWS & BAE-R & Geometry & TextFooler & PWWS & BAE-R\\
\midrule
\midrule

\multirow{4}*{\textbf{AG}} & - & \np & $94.2$ & $86.2$ & $87.6$ & $63.6$ & $17.9$ & $18.6$ & $25.7$ & $20.9$ & $7.4$ 
\\  \cmidrule(lr){2-3} \cmidrule(lr){4-4} \cmidrule(lr){5-8} \cmidrule(lr){9-12}
& \multirow{2}*{\ind} & \btcl & $94.4$ & $80.6$ & $84.6$ & $63.1$ & $17.7$ & $18.1$ & $24.6$ & $20.9$ & $\textbf{7.5}$
\\  & & \adcl & $94.3$ & $\textbf{76.9}$ & $\textbf{80.7}$ & $\textbf{55.9}$ & $\textbf{14.1}$ & $\textbf{19.1}$ & $\textbf{26.7}$ & $\textbf{22.6}$ & $\textbf{7.5}$ 
\\\cmidrule(lr){2-3} \cmidrule(lr){4-4} \cmidrule(lr){5-8} \cmidrule(lr){9-12}
  & \out & \adcl & $94.1$ & $\textit{79.2}$ & $\textit{84.0}$ & $\textit{60.4}$ & $\textit{16.3}$ & $\textit{18.7}$ & $\textit{25.9}$ & $\textit{21.9}$ & $\textbf{7.5}$\\
\midrule

\multirow{4}*{\textbf{IMDB}} & - & \np & $92.3$ & $98.7$ & $99.0$ & $99.2$ & $54.0$ & $3.5$ & $6.5$ & $4.3$ & $3.0$
\\  \cmidrule(lr){2-3} \cmidrule(lr){4-4} \cmidrule(lr){5-8} \cmidrule(lr){9-12}
 & \multirow{2}*{\ind}& \btcl & $92.5$ & $93.3$ & $96.6$ & $95.1$ & $52.0$ & $\textbf{4.5}$ & $7.4$ & $4.4$ & $\textbf{3.3}$ 
\\ & & \adcl & $92.4$ & $\textbf{84.2}$ & $\textbf{87.8}$ & $\textbf{87.8}$ & $\textbf{48.0}$ & $3.7$ & $\textbf{8.7}$ & $\textit{5.1}$ & $2.3$
\\\cmidrule(lr){2-3} \cmidrule(lr){4-4} \cmidrule(lr){5-8} \cmidrule(lr){9-12}
 & \out & \adcl & $92.5$ & $\textit{92.3}$ & $\textit{95.7}$ & $\textit{94.5}$ & $\textit{50.1}$ & $\textit{4.4}$ & $\textit{8.6}$ & $\textbf{5.3}$ & $\textit{3.1}$\\

\bottomrule
\end{tabular}
}

\caption{\label{tab:ood}
Comparison of \out with \ind. We use the DBpedia dataset as the out-of-domain dataset for AG's News and IMDB. Models are finetuned on clean examples after pretraining (\ftc). Best results are \textbf{bolded}, while the second best are in \textit{italic}.
}
\end{table*}

\noindent \textbf{H2}: To validate H2, we compare two settings of $\np+\adv$ and $\adcl+\adv$. We note that when compared with conducting adversarial training alone ($\np+\adv$), \textit{combining our self-supervised contrastive learning method with adversarial training ($\adcl+\adv$) constantly results in higher robustness}. In other words, the adversarial attacks have lower success rates and higher replacement rate in $\adcl+\adv$ models than in $\np+\adv$ models. For instance, for the IMDB dataset, the $\adcl+\adv$ model is 2.7\% more robust than the $\np+\adv$ model, when both models are tested against the Geometry Attack (Success rates of Geometry attack: $\adcl+\adv$: 48.7\%, $\np+\adv$: 51.4\%; Replacement rates: $\adcl+\adv$: 8.1\%, $\np+\adv$: 7.4\%). 

Note that when test $\np+\adv$ models and $\adcl+\adv$ models against the other three adversarial attacks, $\adcl+\adv$ models do not show an advantage over $\np+\adv$ models in terms of replacement rates, despite that $\adcl+\adv$ models still constantly make lower success rates against the adversarial attacks. We argue that this is because we use the Geometry Attack for adversarial training during finetuning, and the adversarial examples from the Geometry Attack might not fully match the distribution from the other attacks. Nevertheless, we can still conclude that $\adcl+\adv$ models are more robust than $\np+\adv$ models.

Our experiments also show that during contrastive learning, the queue size (see~\cref{sec:adv_cl}) has an impact on the final performance. We give the detailed analysis in~\cref{app:queue}.

\begin{table}[!t]
\renewcommand\arraystretch{1.0}
\centering
\resizebox{0.8\linewidth}{!}{%

\begin{tabular}{cccc}
\toprule
\multirow{2}*{\textbf{Dataset}} & \multirow{2}*{\textbf{Attack}} & \multicolumn{2}{c}{\textbf{Success Rate (\%)}} \\ \cline{3-4} & & \np $\rightarrow$ \adcl & \adcl $\rightarrow$ \np\\
\midrule
\midrule
\multirow{4}*{\textbf{AG}} & Geometry & $30.2$ & $62.3$ \\ & TextFooler & $19.7$ & $55.0$  \\ & BAE-R & $26.4$ & $60.2$  \\   & PWWS & $28.3$ & $50.4$  \\
\midrule
\multirow{4}*{\textbf{Yelp}} & Geometry &$30.1$ & $36.4$ \\ & TextFooler & $22.4$ & $28.0$ \\ & BAE-R & $37.4$ & $41.5$  \\   & PWWS & $34.8$ & $36.3$ \\
\midrule
\multirow{4}*{\textbf{IMDB}} & Geometry & $38.2$ & $41.4$ \\ & TextFooler & $22.1$ & $25.2$ \\ & BAE-R & $28.9$ & $30.8$ \\   & PWWS & $24.7$ & $26.0$ \\
\midrule
\multirow{4}*{\textbf{DBpedia}} & Geometry & $34.6$ & $52.2$ \\ & TextFooler & $27.5$ & $42.8$ \\ & BAE-R & $32.5$ & $55.8$ \\   & PWWS & $55.3$ & $58.8$ \\
\bottomrule
\end{tabular}
}

\caption{\label{tab:transfer}
Transferability of adversarial examples. The models are pretrained under either \np or \adcl, and then finetuned on clean examples. \np $\rightarrow$ \adcl: Generate adversarial examples with the model pretrained with \np, then test the model pretrained with \adcl on these adversarial examples. Same applies to \adcl $\rightarrow$ \np.
}
\end{table}

\smallskip

\begin{figure*}
    \centering
    \subfigure[$\np+\ftc$]{\includegraphics[width=0.24\linewidth]{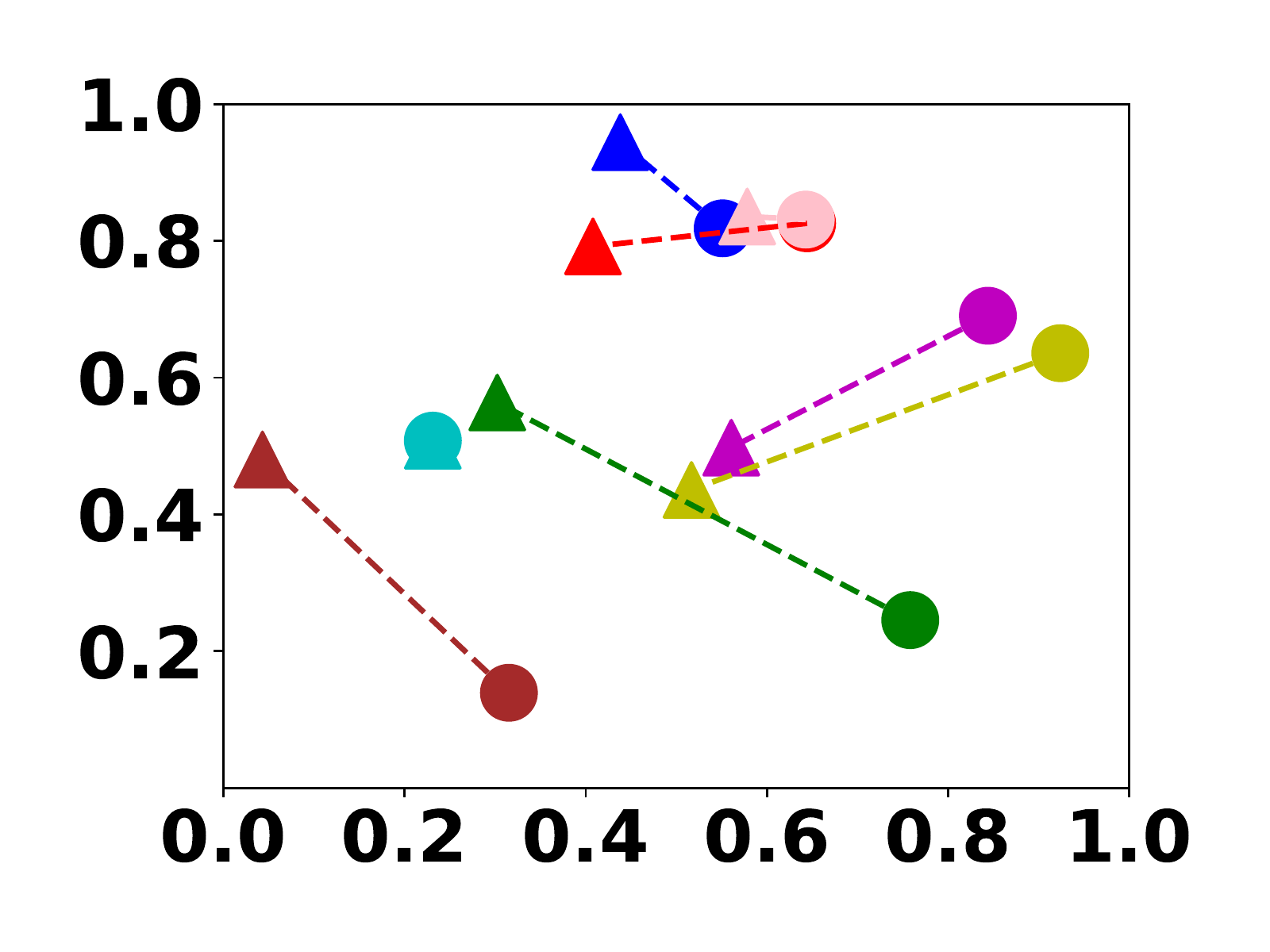}} 
    \subfigure[$\adcl+\ftc$]{\includegraphics[width=0.24\linewidth]{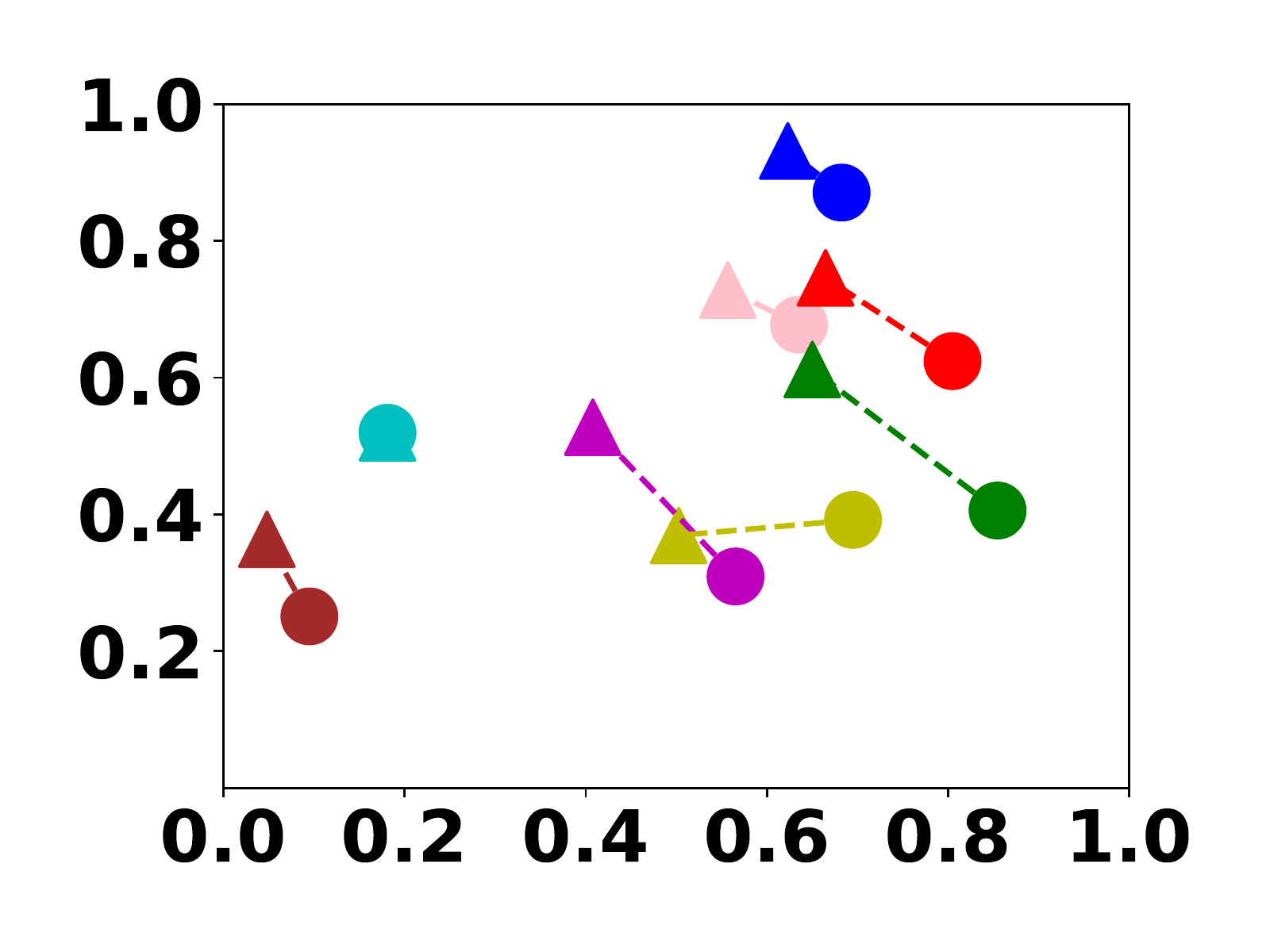}} 
    \subfigure[$\np+\adv$]{\includegraphics[width=0.24\linewidth]{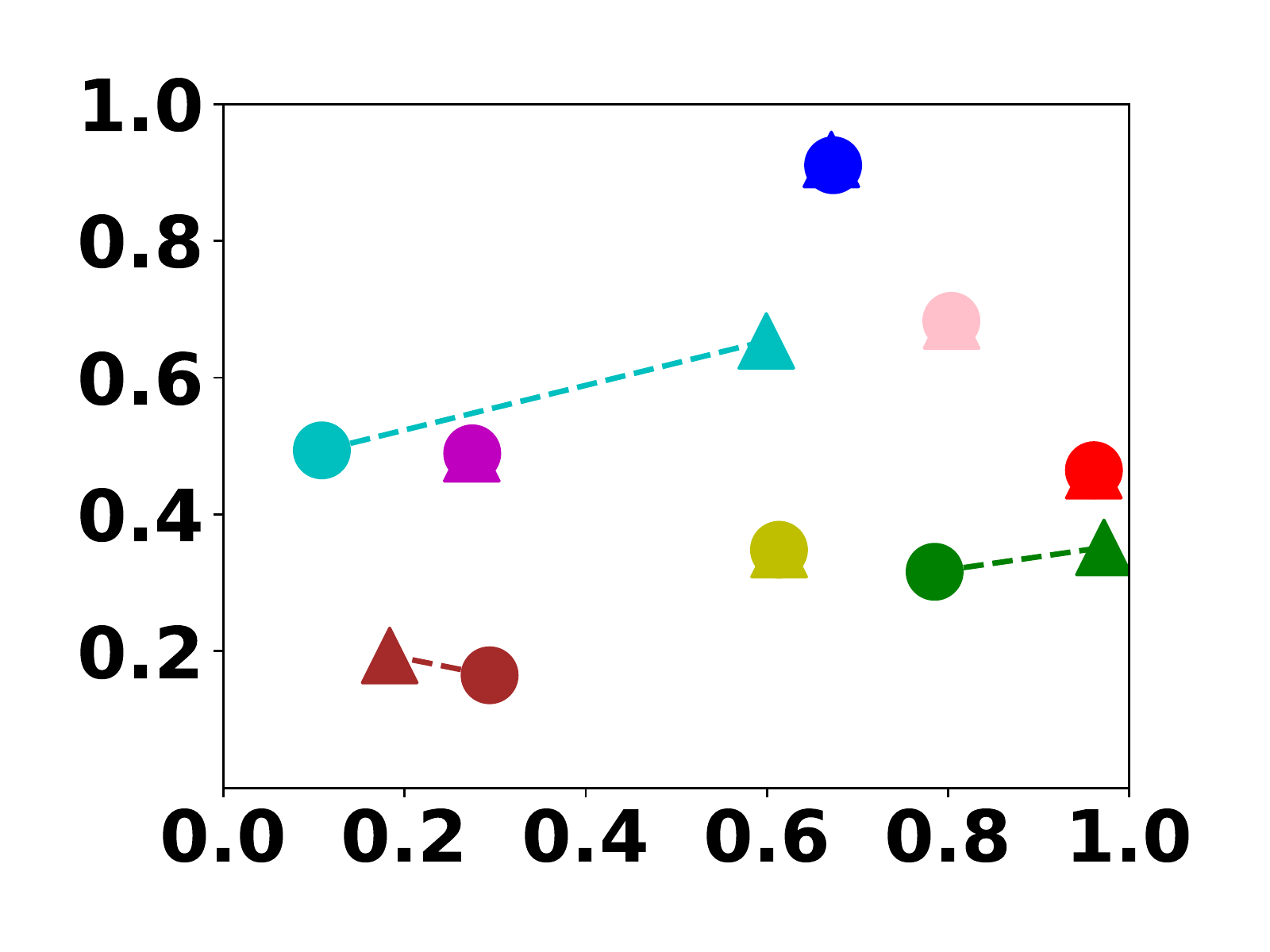}} 
    \subfigure[$\adcl+\adv$]{\includegraphics[width=0.24\linewidth]{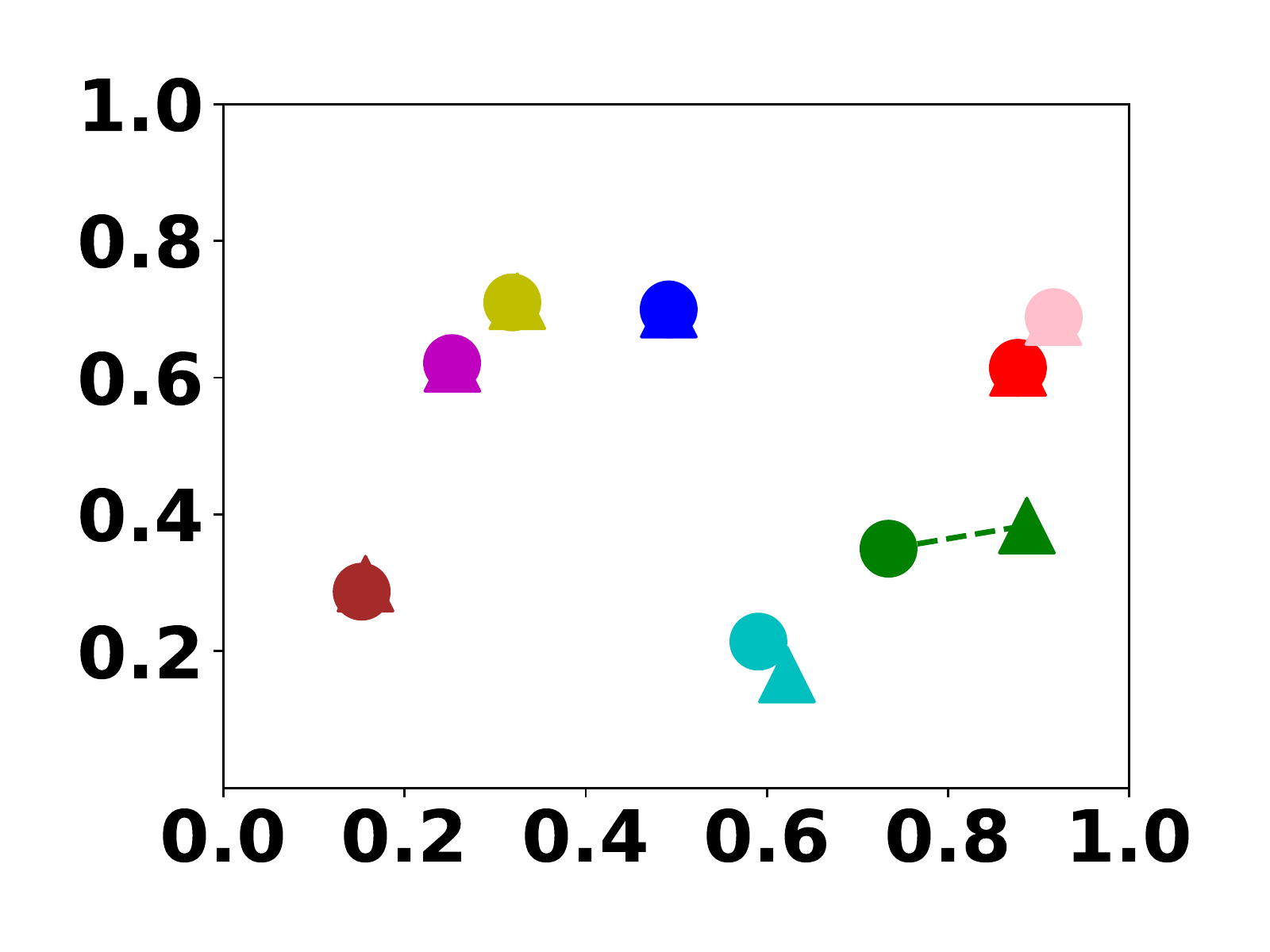}} 
    \caption{t-SNE plot of the vector representations of clean examples and adversarial examples from the AG's News dataset. Markers of the same color indicate a pair of clean example ($\circ$) and adversarial example ($\small \triangle$). Check~\cref{sec:results} for the evaluation settings. The ranges of x-axis and y-axis are normalized to [0, 1]. We connect each clean example by a dotted line to its corresponding adversarial example. }
    \label{fig:vec_ag}
\end{figure*}

\noindent \textbf{H3}:
For the \out setting, we use the DBpedia dataset as the out-of-domain dataset for the AG's News and IMDB datasets, mainly because (1) Computational limits: While using larger datasets such as BookCorpus or Wikipedia might be more useful, conducting self-supervised contrastive learning on these datasets exceeds the limits of our computational infrastructure; (2) The DBpedia dataset is several times larger than AG's News and IMDB. This should give us a glimpse of what it looks like when we scale self-supervised contrastive learning with adversarial perturbations to even larger out-of-domain datasets; (3) The DBpedia dataset (topic classification on Wikipedia) has a different task and domain compared to the AG's News dataset (news classification from a newspaper) and IMDB dataset (sentiment classification on movie reviews). This discrepancy allows us to understand how \textit{out-of-domain} datasets could help.

Table~\ref{tab:ood} shows our results. We can see that models pretrained with \adcl under the \out setting are more robust than models without any pretraining at all (\np). This shows that \textit{our method can improve model robustness using out-of-domain data}. For instance, for the IMDB dataset, the success rate of TextFooler decreases from 98.7\% for FT models to 92.3\% for \out \adcl models. This shows that our method can improve the model robustness even if the dataset used for contrastive learning is from a completely different domain. Note that in \cref{tab:ood}, after pretraining, we finetune the model on clean examples (\ftc).

We also notice that models pretrained with \adcl under the \out setting are not as robust as models pretrained with \adcl under the \ind setting. This indicates we might need to use much larger unlabeled raw datasets to obtain more improvements.

\begin{table}[!ht]
\centering
\renewcommand\arraystretch{0.99}
\resizebox{0.95\linewidth}{!}{%
\begin{tabular}{ccccc}
\toprule
\multirow{2}*{\textbf{Dataset}} & \multicolumn{4}{c}{\textbf{Distance ($d_{pos}/d_{neg}/\delta$)}} \\ \cmidrule{2-5} & \multicolumn{1}{c}{$\np+\ftc$} & \multicolumn{1}{c}{$\adcl+\ftc$} &
\multicolumn{1}{c}{$\np+\adv$} & \multicolumn{1}{c}{$\adcl+\adv$} \\
\midrule
\midrule
\textbf{AG} & $2.4 / 3.9 / 1.5$ & $1.8 / 4.0 / 2.2$ & $0.7 / 4.1 / 3.4$ & $0.7 / 4.4 / 3.7$ \\
\midrule
\textbf{Yelp} & $3.5 / 3.7 / 0.2$ & $2.9 / 4.0 / 1.1$ & $0.7 / 3.2 / 2.5$ & $0.5 / 3.4 / 2.9$\\
\midrule
\textbf{IMDB} & $3.0 / 3.7 / 0.7$ & $2.3 / 3.8 / 1.5$ & $0.6 / 3.4 / 2.8$ & $0.6 / 3.8 / 3.2$\\
\midrule
\textbf{DBpedia} & $2.8 / 4.8 / 2.0$ & $2.3 / 5.1 / 2.8$ & $0.4 / 4.9 / 4.5$ & $0.4 / 5.2 / 4.8$\\
\bottomrule
\end{tabular}
}

\caption{\label{tab:distance}
Vector space study. For each setting, we evaluate three metrics: (a) Average distance between positive pairs; (b) Average distance between negative pairs; (c) Difference between (a) and (b).
}
\end{table}

\smallskip
\noindent \textbf{H4}: To validate this hypothesis, we study the vector representations of $M=1000$ clean examples of the AG's News dataset and their corresponding adversarial examples. We obtain the adversarial examples by attacking a $\np+\ftc$ model. 

Let $\bm v_1,\bm v_2 ...\bm v_{M}$ and $\bm v_{1}^\prime,\bm v_{2}^\prime...\bm v_{M}^\prime$ be the vector representations of the clean examples and corresponding adversarial samples, respectively. For each setting, we evaluate three metrics: 
\begin{itemize}[topsep=0pt,itemsep=0pt,partopsep=0pt, parsep=0pt, leftmargin=*] 
\item Average distance $d_{pos}$ between each of the positive pairs $v_i$ and $v_{i}^\prime$, where $1 \leq i \leq M$. Then we have: 

\begin{align*}
    d_{pos} = \frac{1}{M} \sum_{i=1}^{M} d(\bm v_i, \bm v_{i}^\prime)
\end{align*}

\noindent where $d(\cdot, \cdot)$ denotes the distance between two vectors.

\item Average distance $d_{neg}$ between negative pairs: 
\begin{align*}
    d_{neg} = \sum_{i=1}^{M} \sum_{j=1}^{M} \frac{\mathbbm{1}_{i \neq j} ( d(\bm v_i, \bm v_j) + d(\bm v_i, \bm v_{j}^\prime)}{2(M-1)}
\end{align*}
\item Difference $\delta = d_{neg} - d_{pos}$ between (a) and (b). 
\end{itemize}

Furthermore, we evaluate the above metrics under the following settings:
\begin{itemize}[topsep=0pt,itemsep=0pt,partopsep=0pt, parsep=0pt, leftmargin=*] 
    \item {$\np+\ftc$}: Finetune on clean examples.
    \item {$\adcl+\ftc$}: First do \adcl pretraining, and then finetune on clean examples.
    \item {$\np+\adv$}: Finetune with adversarial training.
    \item {$\adcl+\adv$}: First do \adcl pretraining. Then finetune with adversarial training.
\end{itemize}

\noindent \Cref{tab:distance} shows the results. We can see that our method (1) increases the distance between negative pairs in all settings; (2) decreases the distance between positive pairs in $\np+\ftc$ and $\adcl+\ftc$ models, while the distances between positive pairs barely change in $\np+\adv$ and $\adcl+\adv$ models; (3) increases $\delta$ in all settings. The above observations validate H4 in~\cref{sec:exp_designs}, and further explain that our method achieves higher robustness by pushing vector representations of clean examples and adversarial examples closer.

In~\cref{fig:vec_ag}, we further give qualitative analysis on the distances between clean examples and adversarial examples of the AG's News dataset by showing the t-SNE plot. We can see from the plot that the distances between the clean examples and the corresponding adversarial examples are closer when we apply $\adcl$ pretraining, and that combining $\adcl$ with $\adv$ gives the smallest distance between supervised adversarial examples. Additional plots of other datasets are available in~\cref{app:plot}.

\section{Conclusion and Future Work}

In this paper, we improve the robustness of pretrained language models against word substitution-based adversarial attacks by using self-supervised contrastive learning with adversarial perturbations. Our method is different from previous works as we can improve model robustness without accessing annotated labels. Furthermore, we also conduct word-level adversarial training on BERT with \textit{on-the-fly} generated adversarial examples. Our adversarial training is different from previous works in that (1) it is on the word level; (2) we generate adversarial examples on the fly, instead of generating a fixed adversarial set beforehand. 
Experiments show that our method improves model robustness. We find that combining our method with adversarial training results in better robustness than conducting adversarial training alone. In the future, we plan to scale our method to even larger out-of-domain datasets.

\section*{Ethical Considerations}
To the best of our knowledge, the data used in our work does not contain sensitive information.
Although our models are evaluated on academic datasets in this paper, they could also be used in sensitive contexts, e.g. healthcare or legal scenarios. It is essential that necessary anonymization and robustness evaluation is undertaken before using our models in these settings. 

\section*{Acknowledgements}
Mrinmaya Sachan acknowledges support from an ETH Z\"urich Research grant (ETH-19 21-1) and a grant from the Swiss National Science Foundation
(project \#201009) for this work.

\bibliography{anthology}
\bibliographystyle{acl_natbib}

\appendix
\clearpage

\section{Geometry Attack for Contrastive Loss}\label{app:geo}

Algorithm~\ref{alg:attack} is the pseudocode of our Geometry Attack for contrastive loss. Refer to Section~\ref{sec:attack} for more details.

\begin{algorithm*}

\caption{Geometry Attack for Contrastive Loss}
\begin{algorithmic}[1]
\STATE \textbf{Input:} Example $ X_i = \{w_1, w_2, \dots, w_L\}$, encoder $f$ and MLP $g$
\STATE \textbf{Output:} Adversarial example $X_i^\prime$ 
\STATE Initialize $\bm z_i \gets g(f(X_i))$
\FOR {iter = $1$ to $N$}
\STATE calculate $\ell_i$ using Equation~\ref{eq:loss}
\STATE $\bm v_{z_i} \gets \nabla_{\bm z_i} \ell_i$
\STATE $\bm E \gets \texttt{BertEmbeddings}(X_i^\prime) = \{\bm e_1, \bm e_2, \dots, \bm e_L \}$

\STATE $\bm G \gets \nabla_{\bm E} \ell_i = \{\bm g_1, \bm g_2, \dots, \bm g_L \}$
\STATE $t \gets \arg\max_t ||\bm g_t||$
\STATE $C \gets \texttt{BertForMaskedLM}(\{w_1, \cdots, w_{t-1}, \texttt{[MASK]}, w_{t+1}, \cdots, w_L\})$ 
\STATE $C \gets \texttt{Filter}(C)$ // construct candidates set $C = \{w_{t_1}, w_{t_2}, \cdots, w_{t_T}\}$; filter using counter-fitted embeddings
\FOR {each $w_{t_j} \in C, 1 \leq j \leq T$}
\STATE $X_{i_j} \gets {\{w_1, \cdots, w_{t-1}, w_{t_j} , w_{t+1}, \cdots, w_L\}}$
\STATE $\bm z_{i_j} \gets g(f(X_{i_j}))$
\STATE $\bm r_{i_j} \gets \bm z_{i_j} - \bm z_i$
\STATE $\bm p_{i_j} \gets \frac{\bm r_{i_j} \cdot \bm v_{z_i}}{||\bm v_{z_i}||}$
\ENDFOR
\STATE $m \gets \arg\max_j  ||\bm p_{i_j}||$
\STATE $X_i \gets X_{i_m}$
\STATE $\bm z_i \gets \bm z_{i_m}$
\ENDFOR
\STATE $X_i^\prime \gets X_i$
\RETURN $X_i$
\end{algorithmic}
\label{alg:attack}
\end{algorithm*}

\section{Datasets}\label{app:dataset}

\begin{table}[!ht]
\centering
\resizebox{0.7\linewidth}{!}{%

\begin{tabular}{ccccc}
\toprule
\textbf{Dataset} & \textbf{Labels} & \textbf{Avg Len} & \textbf{Train} & \textbf{Test}\\
\midrule
\midrule
AG's News & $4$ & $44$ & $120$K & $7.6$K\\
\midrule
IMDB & $2$ & $292$ & $25$K & $25$K\\
\midrule
DBPedia & $14$ & $67$ & $560$K & $70$K \\
\midrule
Yelp & $2$ & $177$ & $560$K & $38$K \\
\bottomrule
\end{tabular}
}
\caption{\label{tab:statistics}
Statistics of the datasets.
}
\end{table}

The statistics of each dataset are shown in Table~\ref{tab:statistics}. In our work, the maximum sequence length is set to 128 for AG's News and DBpedia, 256 for Yelp, and 512 for IMDB. To save time during evaluating the model robustness against attacks, we randomly select a part of the test examples in each dataset for evaluation. Specifically, we select 1,000 samples from IMDB, 2,000 samples from Yelp, and 5,000 samples from DBpedia. We use all 7,600 samples from the AG's News test set for evaluation. 

\smallskip

\noindent \textbf{AG's News}\footnote{\url{http://groups.di.unipi.it/~gulli/AG_corpus_of_news_articles.html}} Topic classification dataset with four types of news articles: World, Sports, Business and Science/Technology.
\smallskip

\noindent \textbf{IMDB}~\cite{imdb} Binary sentiment classification dataset on positive and negative movie reviews.
\smallskip

\noindent \textbf{Yelp} Yelp review dataset for binary sentiment classification. Following~\citet{zhang2015character}, reviews with star 1 and 2 are considered negative, and reviews with star 3 and 4 are considered positive.

\smallskip

\noindent \textbf{DBpedia}~\cite{zhang2015character} Topic classification dataset with 14 non-overlapping classes. Both content and title fields are used in our work.

\section{Effect of Queue Size}\label{app:queue}

We conduct additional experiments to study the effect of queue size. We use a queue size of 8192, 16384, 32768, and 65536 under the setting of ADCL+FT for the AG's News dataset.
As is shown in Table~\ref{tab:batchsize}, a larger queue size generally helps improve the model robustness. However, we also notice that when the queue size is too large (65536), the model robustness starts to decrease. We argue that this is because a too large queue size results in less frequent queue updates, which makes the vectors in the queue stale. 

\begin{table}
\centering
\resizebox{0.95\linewidth}{!}{%

\begin{tabular}{cccc}
\toprule
\textbf{Queue Size} & \textbf{Original Acc. (\%)} & \textbf{Success (\%)} & \textbf{Replaced (\%)} \\
\midrule
\midrule
Vanilla & $94.2$ & $86.2$ & $18.6$ \\
\midrule
8192 & $94.4$ & $77.8$ & $18.9$ \\
\midrule
16384 & $94.3$ & $76.9$ & $18.7$ \\
\midrule
32768 & $94.3$ & $\textbf{76.5}$ & $19.1$ \\
\midrule
65536 & $94.4$ & $ 76.7$ & $19.3$ \\
\bottomrule
\end{tabular}
}
\caption{\label{tab:batchsize}
Effect of queue size. We use the Geometry Attack to evaluate the robustness of each model. The FT model is finetuned without contrastive learning.
}

\end{table}

\section{Speed of Different Attacks}\label{app:speed}

We show in Table~\ref{tab:speed} the average number of seconds each attack needs for one example. We obtain the average time by attacking 1000 examples and then taking the average. We can observe that the Geometry attack is at least four times faster than TextFooler, and 4 to 10 times faster than PWWS and BAE-R. 

\begin{table}[!ht]
\centering
\resizebox{0.83\linewidth}{!}{%

\begin{tabular}{ccccc}
\toprule
\textbf{Attack} & \textbf{AG's News} & \textbf{IMDB} & \textbf{DBpedia} & \textbf{Yelp}\\
\midrule
\midrule
Geometry & $0.44$ & $2.02$ & $0.69$ & $1.16$\\
\midrule
TextFooler & $2.48$ & $8.69$ & $2.89$ & $4.86$\\
\midrule
PWWS & $6.29$ & $21.86$ & $2.52$ & $10.27$ \\
\midrule
BAE-R & $5.37$ & $24.10$ & $7.74$ & $16.03$ \\
\bottomrule
\end{tabular}
}
\caption{\label{tab:speed}
Average number of seconds each attack needs for an example.
}
\end{table}

\section{Adversarial Training with Pre-generated Examples}\label{app:pre}

We compare two different methods for adversarial training:

\begin{itemize}
    \item \textbf{Pre-generated} We pre-generate for each example in the training set an adversarial example. We then augment the original training set with the adversarial examples. Finally, the model is finetuned on the augmented dataset.
    
    \item \textbf{On-the-fly} This setting is the same as ADV in Table~\ref{tab:white-box}, where we generate adverarial examples on the fly for each mini-batch during training.
\end{itemize}

Table~\ref{tab:otf} shows the results on the AG's News dataset. We can see that on all four attacks, adversarial training with on-the-fly generated adversarial examples gives higher robustness than adversarial training with pre-generated adversarial examples.

\begin{table}[]
\centering
\resizebox{0.99\linewidth}{!}{%

\begin{tabular}{ccccc}
\toprule 
\multirow{2}*{\textbf{Dataset}} & \multicolumn{4}{c}{\textbf{Success Rate / Replaced (\%)}}\\
                                    & Geometry  & TextFooler & PWWS      & BAE-R    \\
\midrule
\midrule
Pre-generated & 55.3/17.1 & 59.4/22.6  & 42.0/17.4 & 16.5/7.3 \\
\midrule
On-the-fly    & 20.7/20.5 & 25.1/29.3  & 26.1/22.3 & 10.7/7.7 \\
\bottomrule
\end{tabular}
}
\caption{\label{tab:otf} Comparison between adversarial training with pre-generated adversarial examples and on-the-fly generated adversarial examples.}
\end{table}

\section{Implementation Details}

In our paper, we use PyTorch Lightning\footnote{\url{https://www.pytorchlightning.ai/}} and HuggingFace Transformers\footnote{\url{https://huggingface.co/transformers/}} in our implementation. We use BERT as the encoder $f(\cdot)$, and the representation of the \texttt{[CLS]} symbol in the last layer is used for $\bm h$. $g(\cdot)$ is a two-layer MLP, of which the output size $c$ is 128. $g(\cdot)$ uses \texttt{Tanh} as activation function in the output layer. We use FP16 in training step to reduce GPU memory usage, and use FusedAdam from DeepSpeed\footnote{\url{https://www.deepspeed.ai/}} as the optimizer. We enable DeepSpeed ZeRO Stage 2 to further speed up training. We conduct all our experiments on 8 RTX TITAN GPUs.

\smallskip
\noindent \textbf{Contrastive learning} For Geometry Attack for contrastive loss, to reach a balance between attack success rate and efficiency, the maximum number of iterations $K$ is set to 10 for AG's News, DBpedia, and Yelp, and 15 for IMDB dataset. We do not perturb words that were already perturbed in previous iterations. For an example $X_i = \{w_1, w_2, \dots, w_L\}$, at most $min\{K, 0.2\cdot L\}$ words can be perturbed. For each word $w_t, 1\leq t\leq L$, the upper limit of the candidate set size $T$ is set to $25$. Due to the various maximum lengths in downstream datasets and GPU memory limits, we use different batch sizes for different datasets. During contrastive learning, the batch size is set to 1024 for AG's News and DBpedia, 448 for Yelp, and 192 for IMDB. 

\smallskip

\noindent \textbf{Fine-tuning} During finetuning, we train the model for two epochs for AG's News and DBpedia, 3 for Yelp, and 4 for IMDB. The learning rate is set to $2e-5$ and is adjusted using linear scheduling.

\smallskip

\noindent \textbf{Adversarial training} For adversarial training, the number of training epochs is set to 3 with an additional first epoch of finetuning on clean examples. The adversarial examples are generated \textit{on the fly} in each batch during training. For the Geometry Attack in adversarial training, at most $\texttt{min}\{K, 0.4\cdot \texttt{len}(X_i)\}$ words can be perturbed in an example. The upper limit of the candidate set size is set to $50$.

\noindent \textbf{Back Translation} We use pretrained translation models \texttt{opus-mt-en-roa} and \texttt{opus-mt-roa-en} from Helsinki-NLP to generate one translation for each example.

\section{Hard Positive Examples from Geometry Attack for Contrastive Loss}
In Table~\ref{tab:examples}, we show hard positive examples generated by our Geometry Attack for contrastive loss from the AG's News dataset.

\begin{table}[!h]
\centering
\renewcommand\arraystretch{1.1}
\resizebox{1.0\linewidth}{!}{%
\begin{tabular}{cp{1.1\columnwidth}}

\toprule
\texttt{Original} & Zurich employees plead guilty in probe new york (reuters) - two senior insurance \textcolor{blue}{underwriters} at zurich american insurance co \textcolor{blue}{pleaded} guilty on tuesday to \textcolor{blue}{misdemeanors} related to bid-rigging in the insurance market. \\ \cline{2-2}

\texttt{Adversarial} & Zurich employees plead guilty in probe new york (reuters) - two senior insurance \textcolor{red}{agents} at zurich american insurance co \textcolor{red}{testified} guilty on tuesday to \textcolor{red}{violations} related to bid-rigging in the insurance market. \\

\toprule

\texttt{Original} & Black watch troops move into \textcolor{blue}{position} the first units of a black watch \textcolor{blue}{battlegroup} are due to arrive today in their new positions south of baghdad as tony blair indicated that more british troops may replace them in the american - controlled zone before the end of the year. \\ \cline{2-2}

\texttt{Adversarial} & Black watch troops move into \textcolor{red}{place} the first units of a black watch \textcolor{red}{operation} are due to arrive today in their new positions south of baghdad as tony blair indicated that more british troops may replace them in the american - controlled zone before the end of the year. \\ 

\toprule

\end{tabular}
}
\caption{\label{tab:examples} 
Hard positive examples generated by Geometry Attack for contrastive loss. \textcolor{blue}{Blue words} in the original examples are replaced by \textcolor{red}{red words} in the adversarial examples.}
\end{table}

\section{Additional t-SNE plots}~\label{app:plot}

\begin{figure}
    \centering
    \subfigure[$\np+\ftc$]{\includegraphics[width=0.8\linewidth]{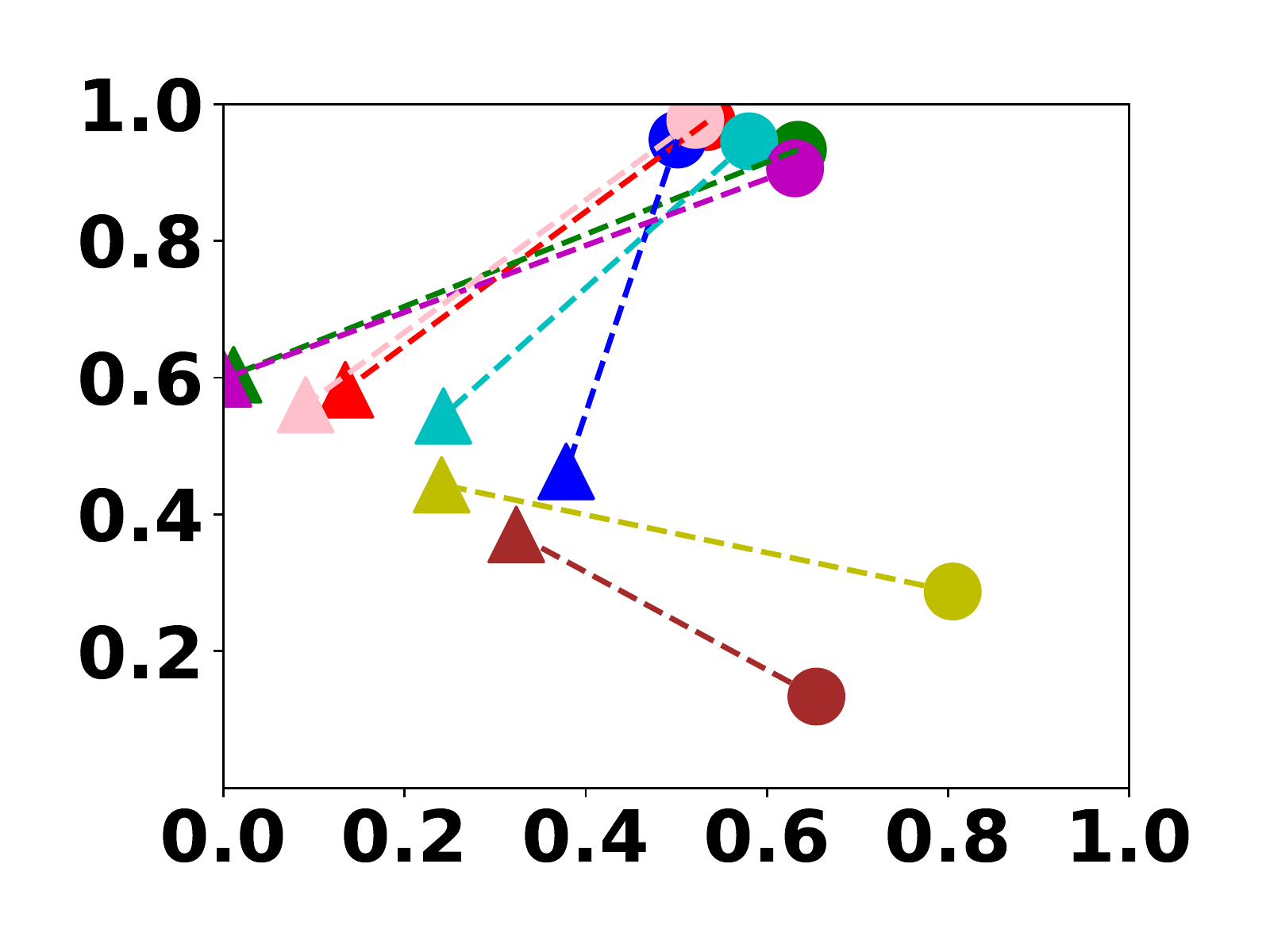}} 
    
    \subfigure[$\adcl+\ftc$]{\includegraphics[width=0.8\linewidth]{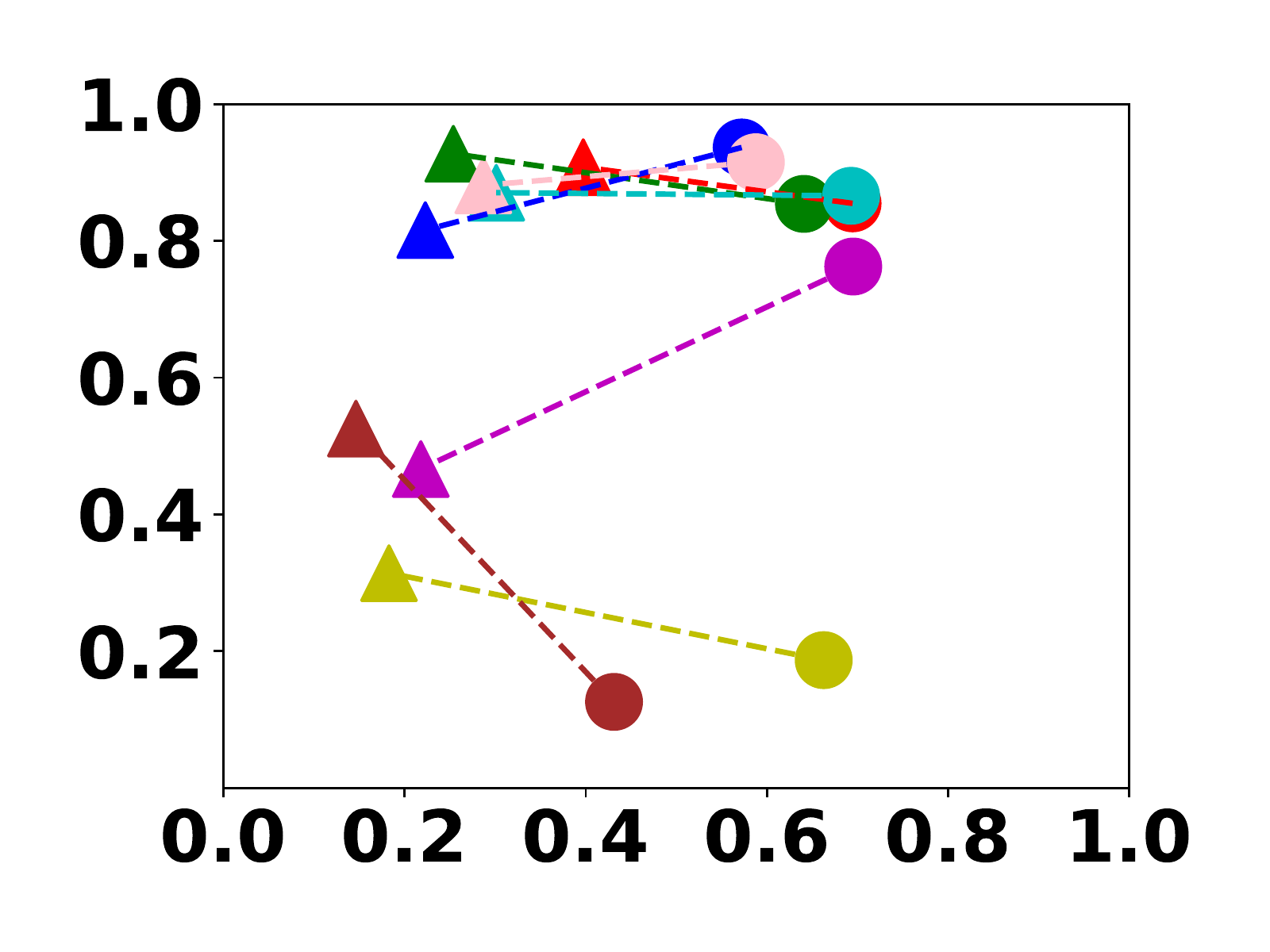}} 
    
    \subfigure[$\np+\adv$]{\includegraphics[width=0.8\linewidth]{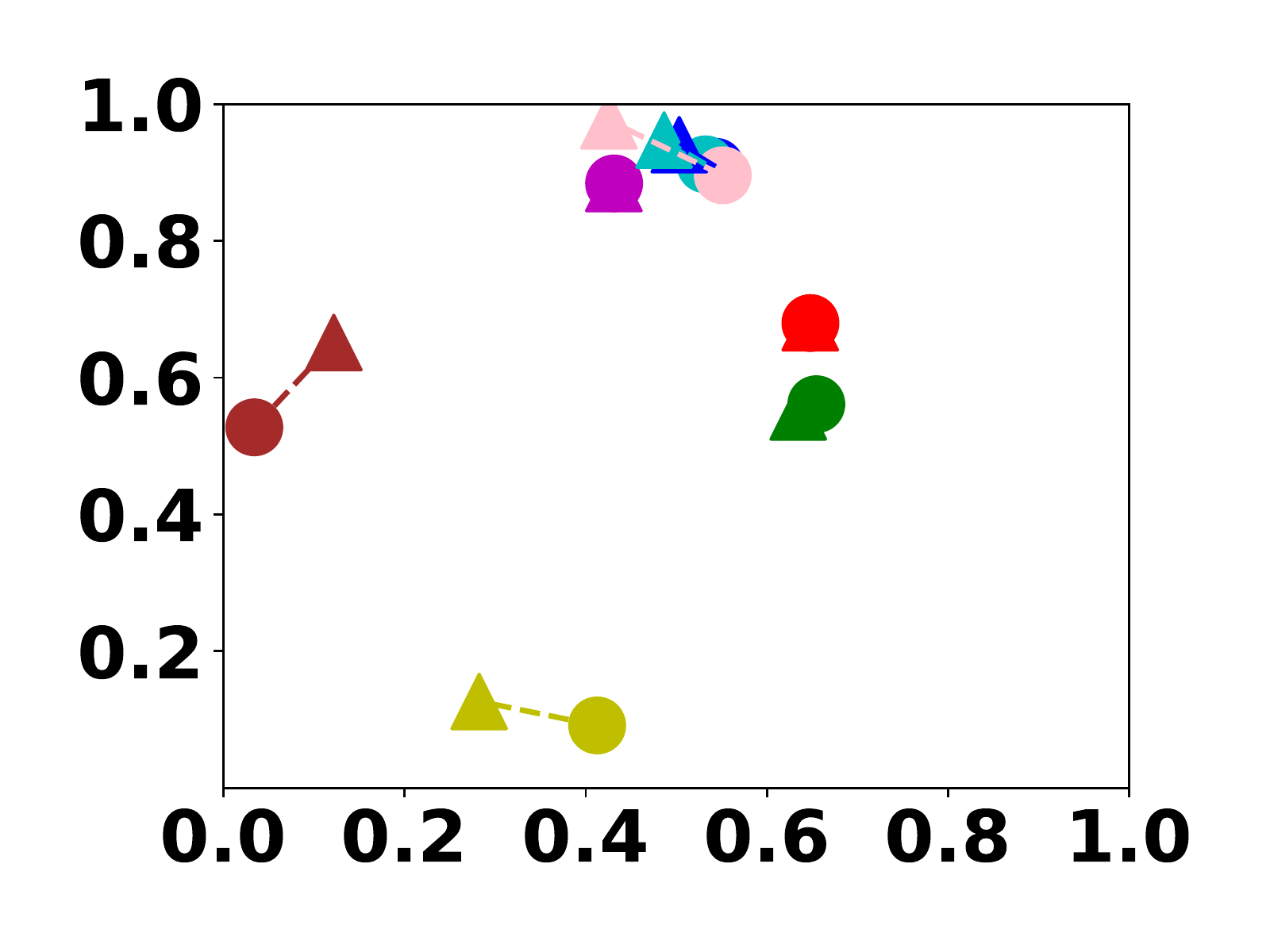}} 
    
    \subfigure[$\adcl+\adv$]{\includegraphics[width=0.8\linewidth]{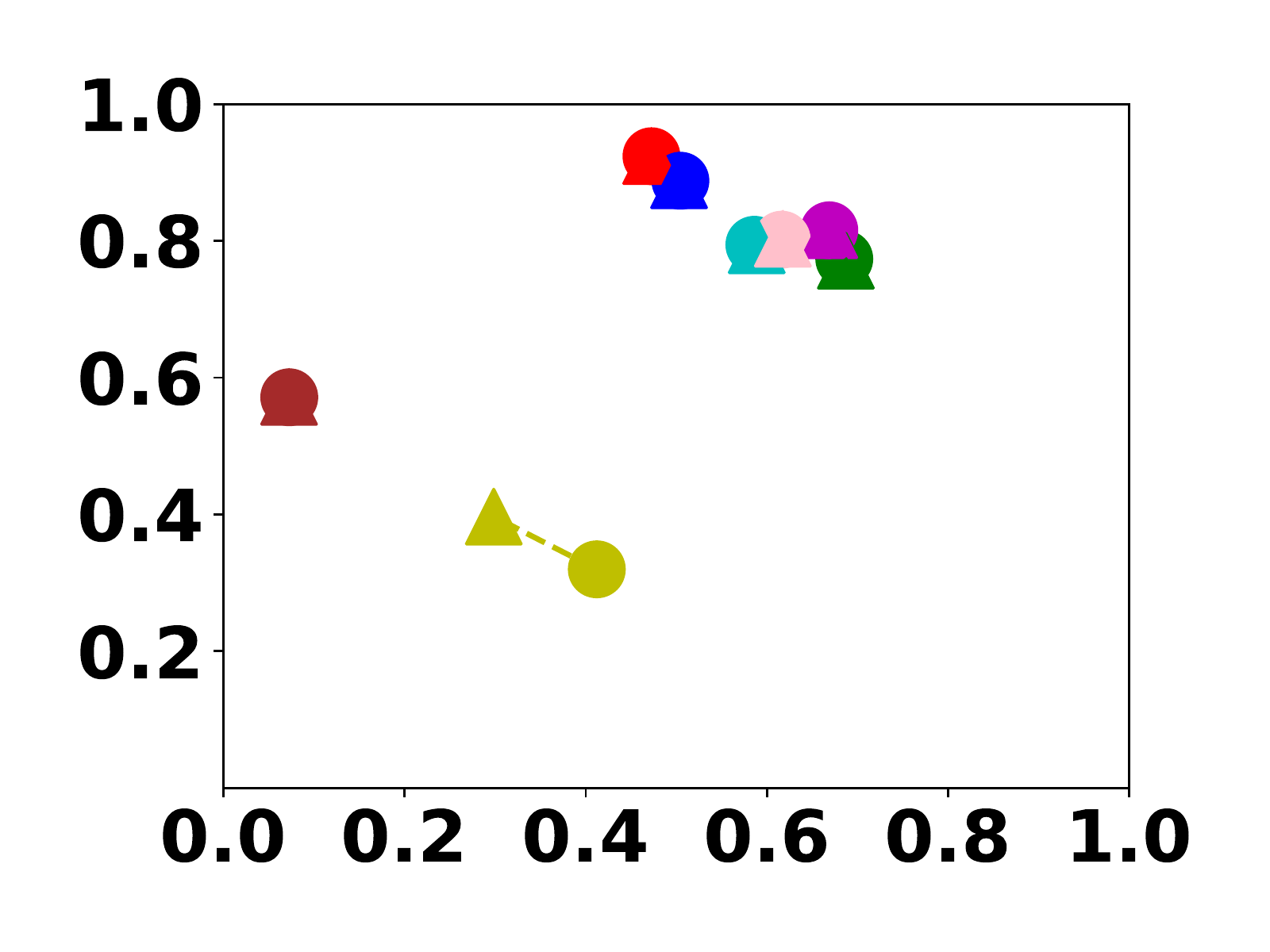}} 

    \caption{t-SNE plot of the vector representations of clean examples and adversarial examples from the Yelp dataset.}
    \label{fig:vec_yelp}
\end{figure}

\begin{figure}
    \centering
    \subfigure[$\np+\ftc$]{\includegraphics[width=0.8\linewidth]{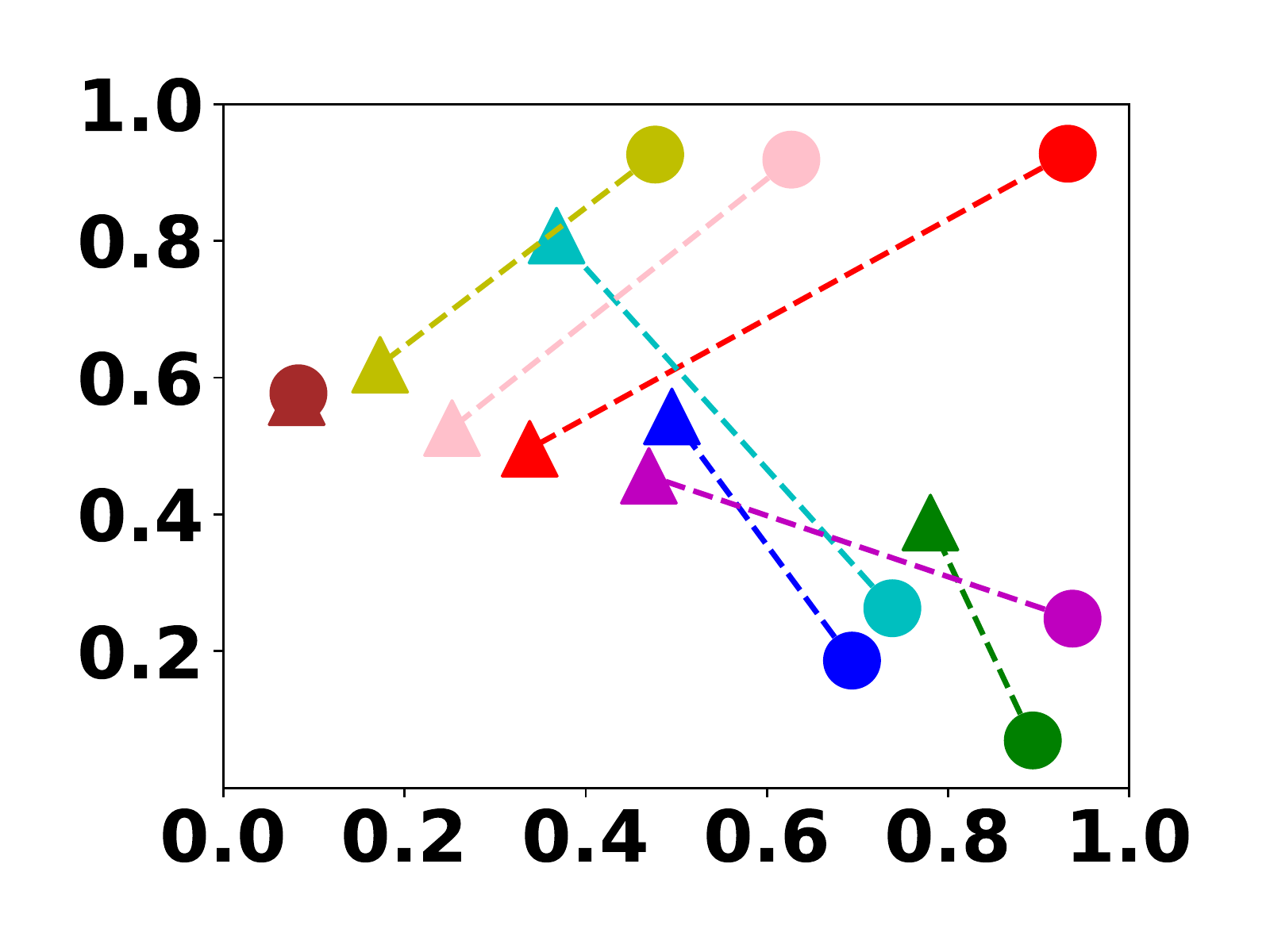}} 
    
    \subfigure[$\adcl+\ftc$]{\includegraphics[width=0.8\linewidth]{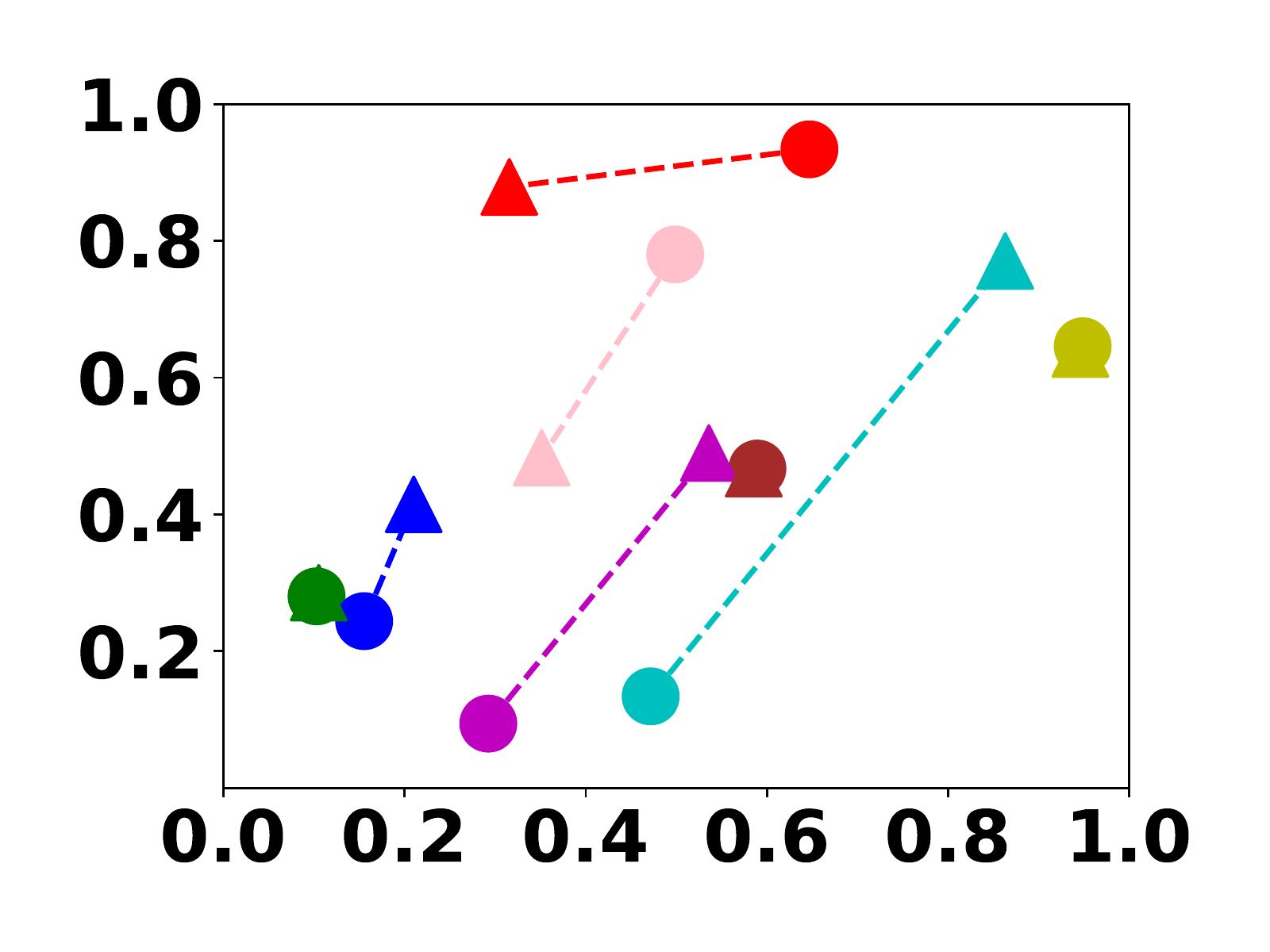}} 
    
    \subfigure[$\np+\adv$]{\includegraphics[width=0.8\linewidth]{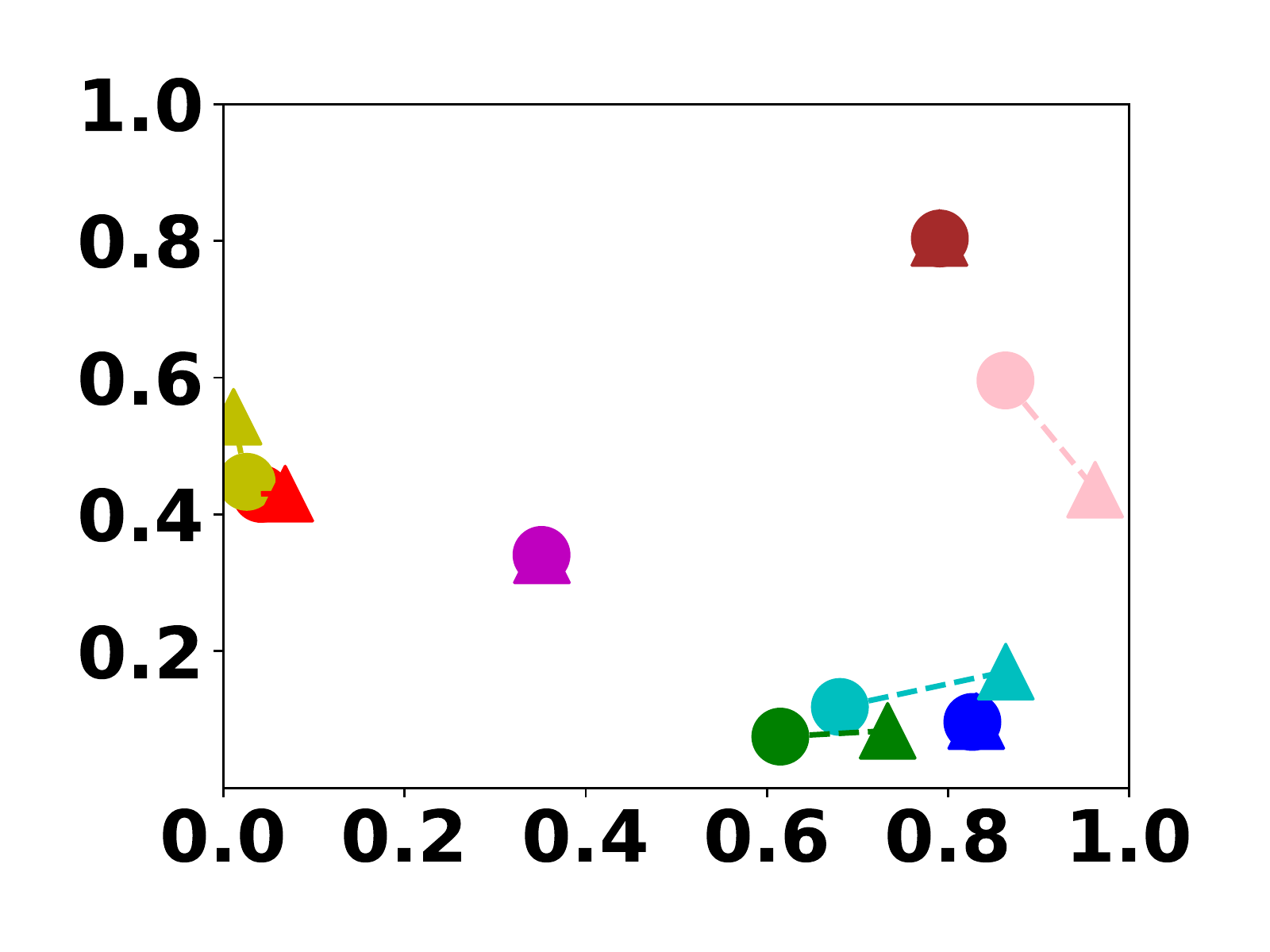}} 
    
    \subfigure[$\adcl+\adv$]{\includegraphics[width=0.8\linewidth]{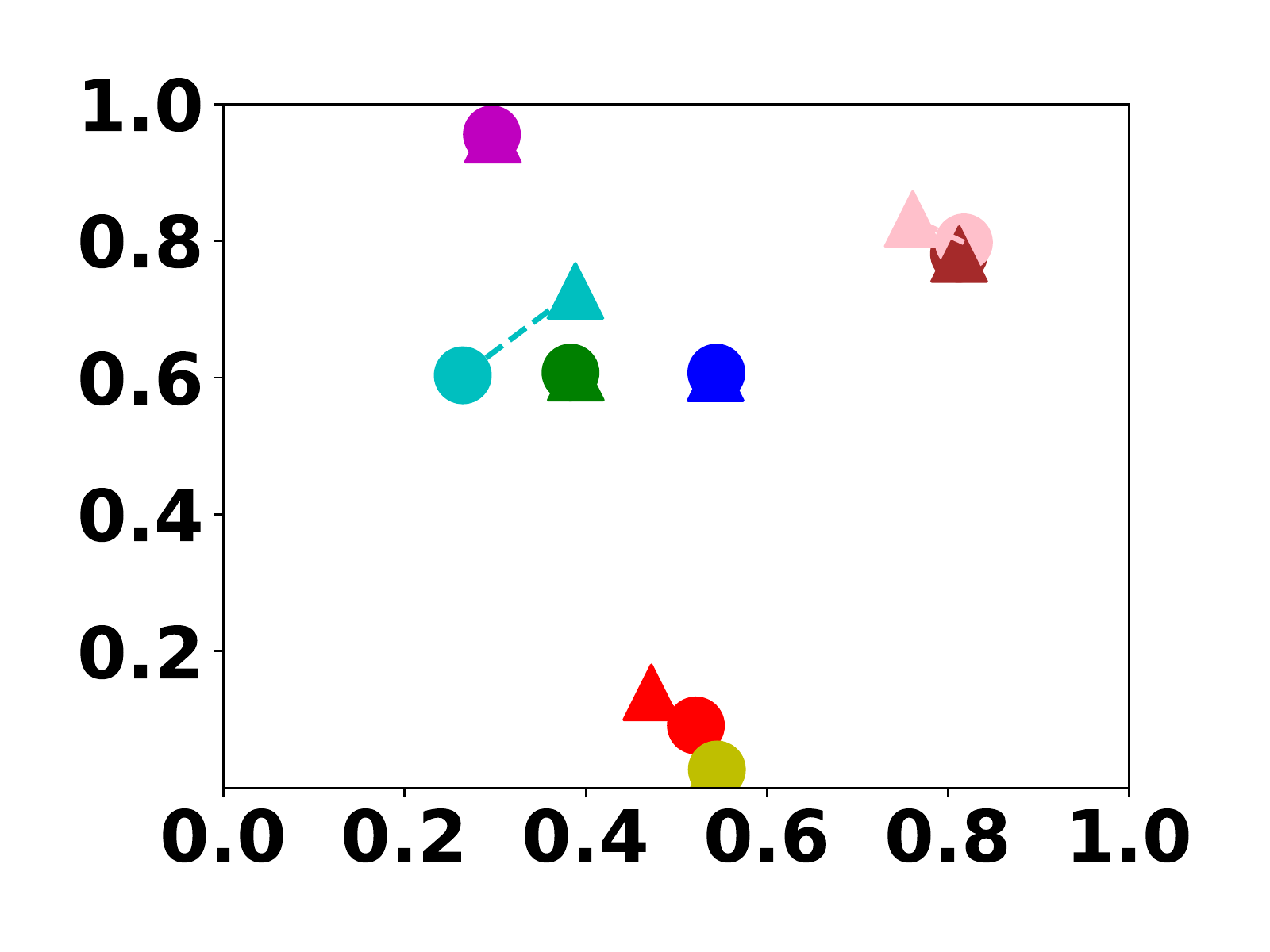}} 

    \caption{t-SNE plot of the vector representations of clean examples and adversarial examples from the IMDB dataset.}
    \label{fig:vec_imdb}
\end{figure}

\begin{figure}
    \centering
    \subfigure[$\np+\ftc$]{\includegraphics[width=0.8\linewidth]{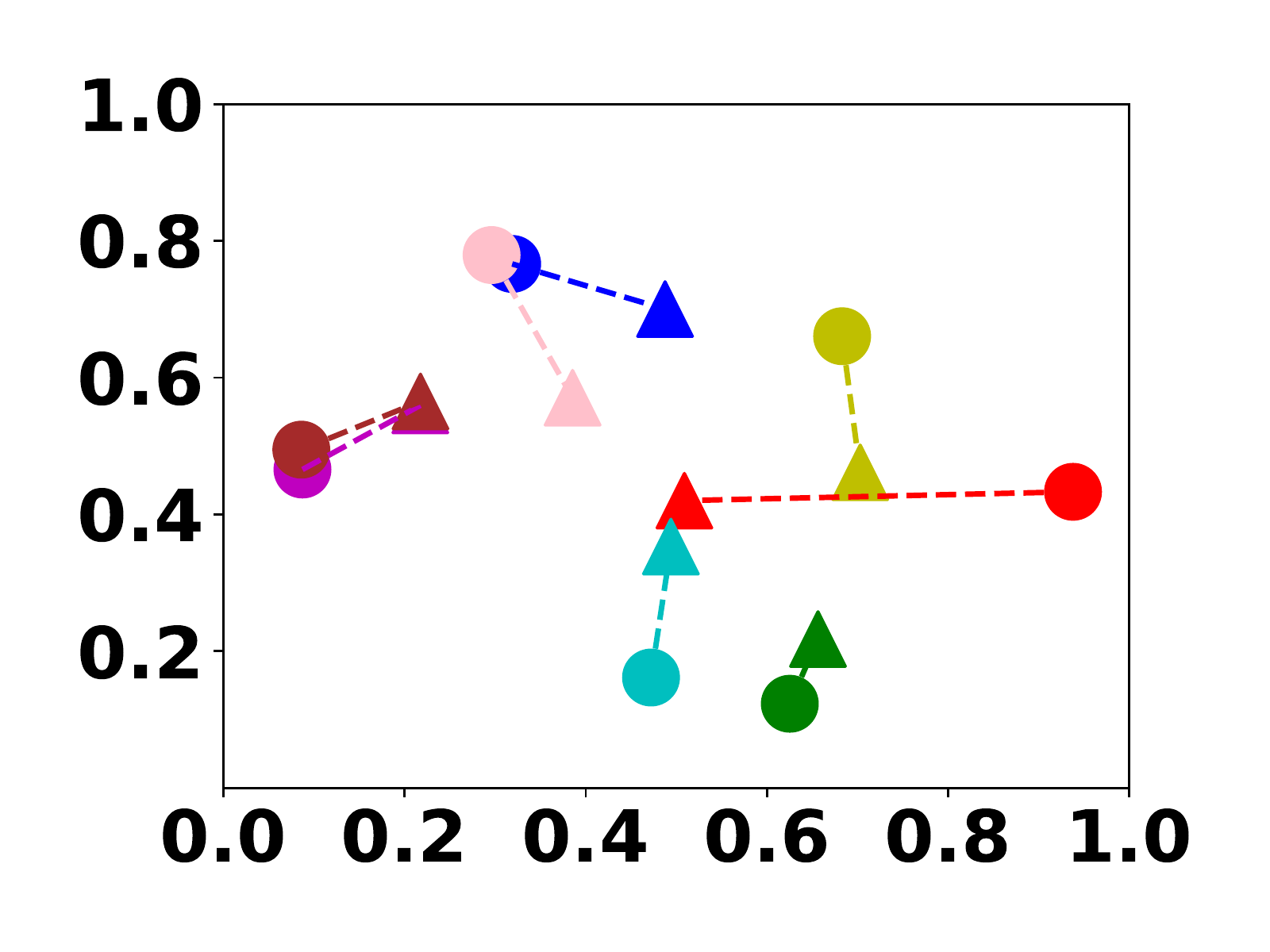}} 
    
    \subfigure[$\adcl+\ftc$]{\includegraphics[width=0.8\linewidth]{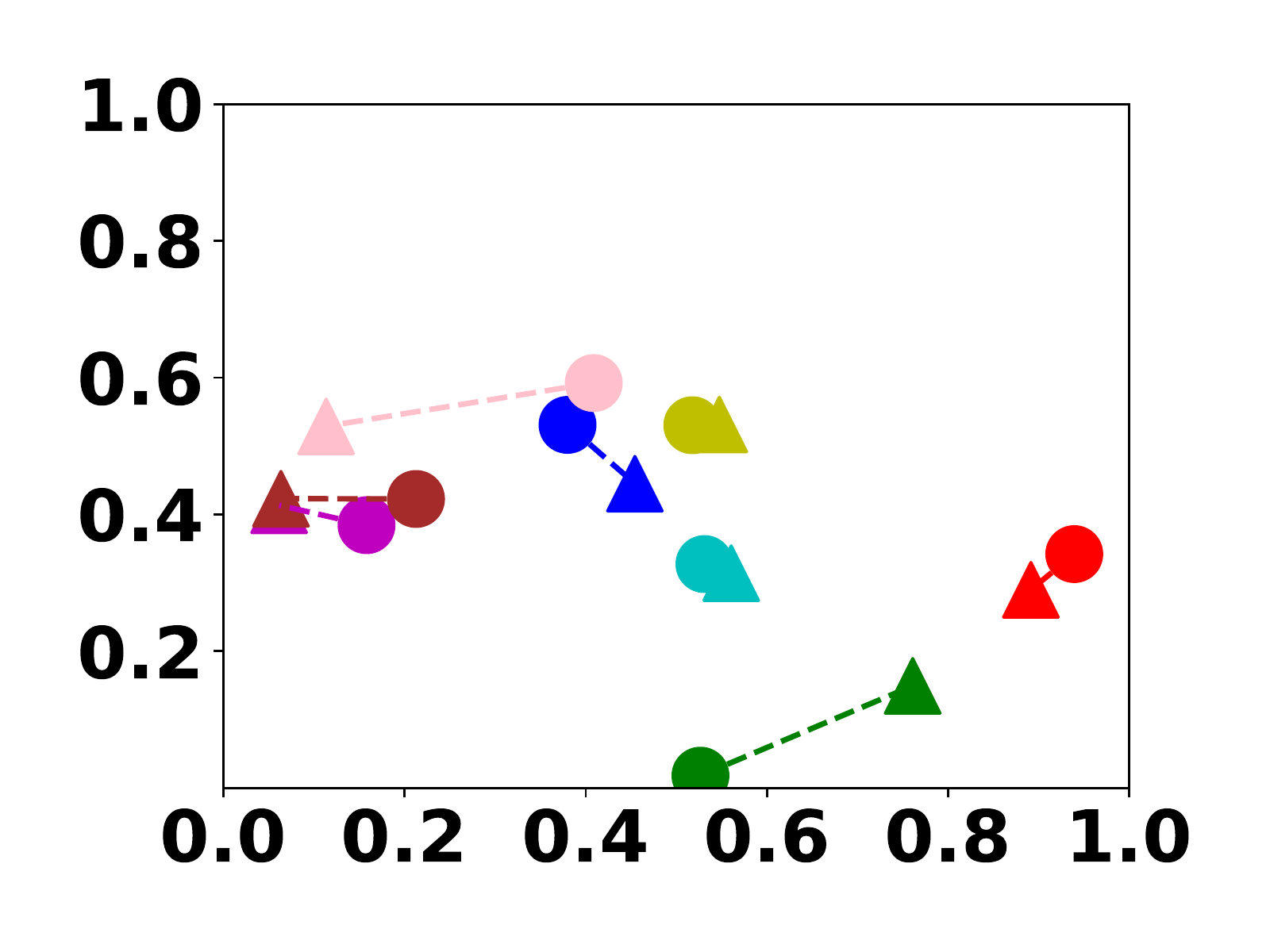}} 
    
    \subfigure[$\np+\adv$]{\includegraphics[width=0.8\linewidth]{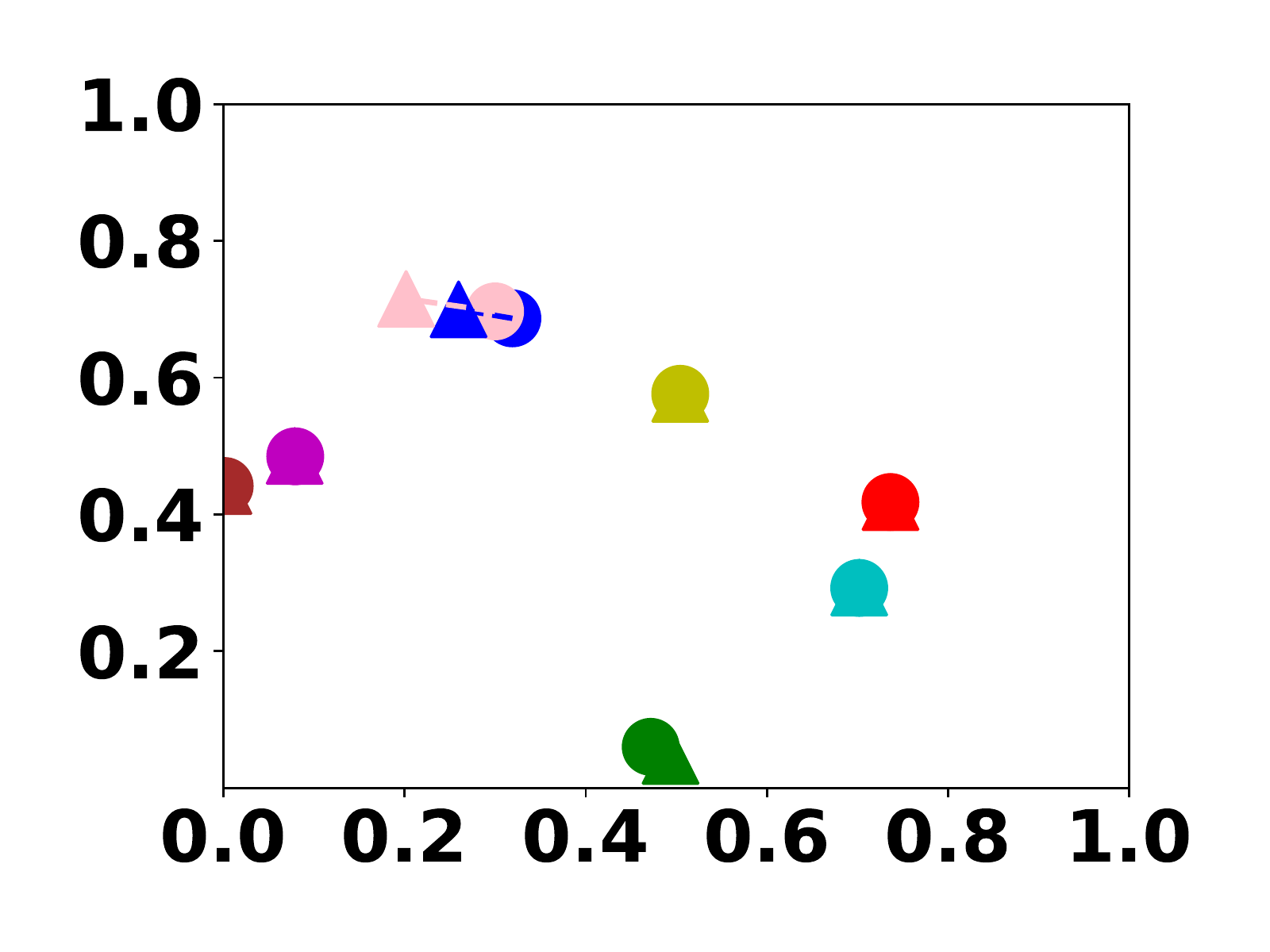}} 
    
    \subfigure[$\adcl+\adv$]{\includegraphics[width=0.8\linewidth]{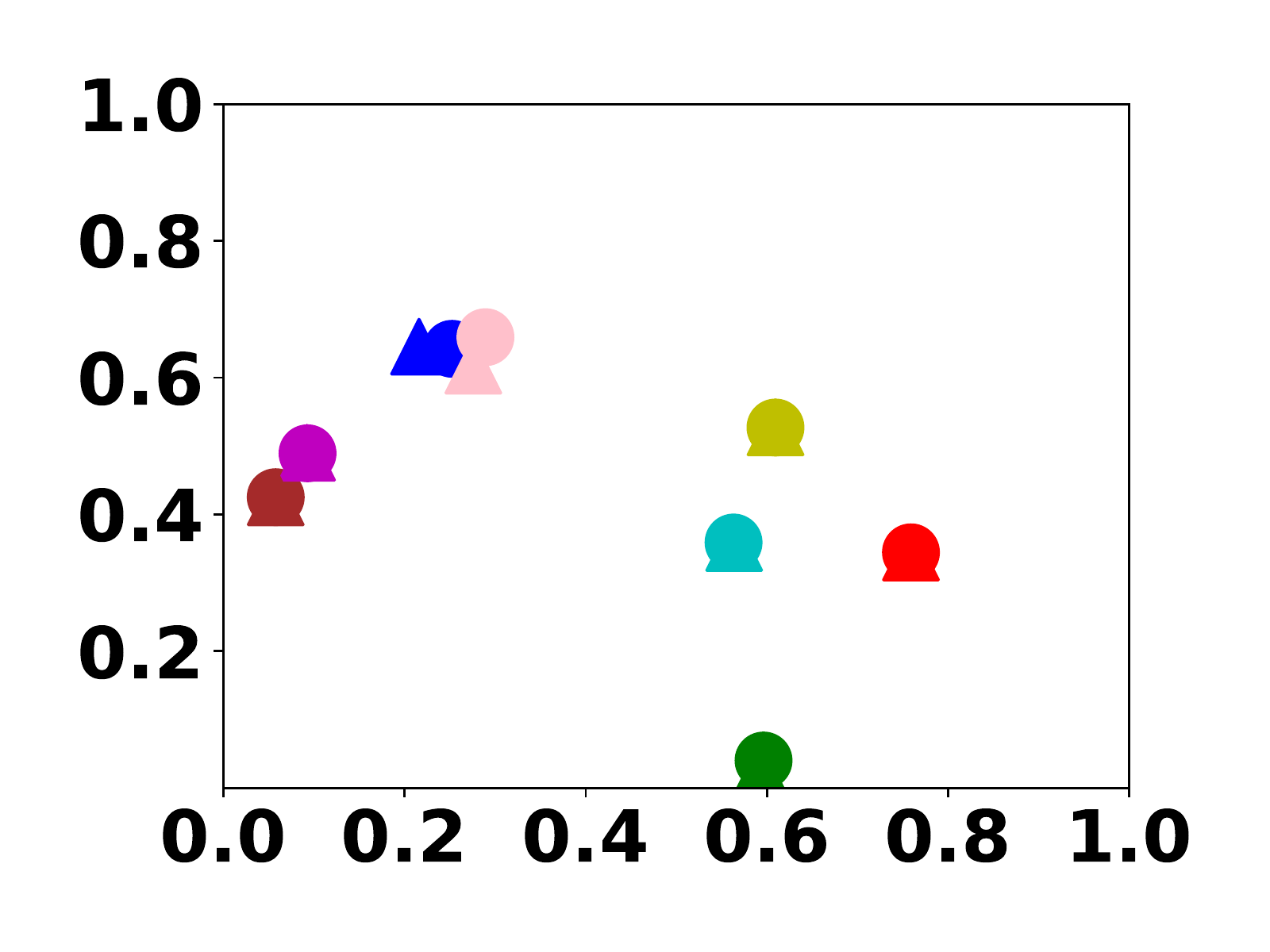}} 

    \caption{t-SNE plot of the vector representations of clean examples and adversarial examples from the DBpedia dataset.}
    \label{fig:vec_dbpedia}
\end{figure}

We give t-SNE plots of the vector representations of clean examples and adversarial examples from Yelp, IMDB and DBpedia in~\cref{fig:vec_yelp},~\cref{fig:vec_imdb} and~\cref{fig:vec_dbpedia}, respectively.

\end{document}